\documentclass[10pt,a4paper]{amsart}
\pdfoutput=1
\usepackage[margin=3.25cm]{geometry}

\usepackage[utf8]{inputenc} 
\usepackage[T1]{fontenc}    
\usepackage{xcolor}         
\usepackage{booktabs}       
\usepackage{amsfonts}       
\usepackage{amsmath}
\usepackage{amsthm}
\usepackage{amssymb}
\usepackage{mathtools}
\usepackage{graphicx}       
\usepackage[shortlabels]{enumitem}
\usepackage[font = small]{caption}
\usepackage{subcaption}     
\usepackage{makecell}       

\makeatletter
\newcommand\thankssymb[1]{\textsuperscript{\@fnsymbol{#1}}}
\makeatother

\usepackage[numbers,sort&compress]{natbib}

\usepackage[colorlinks]{hyperref}       
\hypersetup{linkcolor=blue, citecolor=blue}
\usepackage[nameinlink]{cleveref}       
\usepackage{url}        
\Crefname{equation}{}{}
\Crefname{figure}{Figure}{}

\newcommand{\R}{\mathbb{R}}
\newcommand{\N}{\mathbb{N}}
\DeclareMathOperator{\co}{\overline{co}}
\DeclareMathOperator{\dist}{dist}

\newcommand{\zz}[2]{z_{#1}^{#2}}
\newcommand{\CC}[2]{\mathcal{C}_{#1}(\ZZ^{#2})}

\newcommand{\ZZ}{Z}
\newcommand{\abs}[1]{\left\vert #1 \right\vert}

\newtheorem{theorem}{Theorem}[section]
\newtheorem{corollary}{Corollary}[theorem]
\newtheorem{lemma}[theorem]{Lemma}
\newtheorem{prop}[theorem]{Proposition}
\newtheorem{assumption}[theorem]{Assumption}
\newtheorem{definition}{Definition}[section]
\theoremstyle{remark}
\newtheorem{remark}[theorem]{Remark}
\newtheorem{example}[theorem]{Example}
\numberwithin{equation}{section}

\title[Clustering in pure-attention hardmax transformers]{Clustering in pure-attention hardmax transformers\\and its role in sentiment analysis}

\thanks{Author accepted manuscript. The version of record is published in
\emph{SIAM Journal on Mathematics of Data Science} 7.3 (2025): 1367-1393
\href{https://doi.org/10.1137/24M167086X}{doi:10.1137/24M167086X}}

\author[A. Alcalde]{Albert Alcalde\thankssymb{2}}
\email{albert.alcalde@fau.de}

\author[G. Fantuzzi]{Giovanni Fantuzzi\thankssymb{2}}
\email{giovanni.fantuzzi@fau.de}

\author[E. Zuazua]{Enrique Zuazua\thankssymb{1}\thankssymb{2}\thankssymb{3}}
\email{enrique.zuazua@fau.de}

\thanks{\thankssymb{2}Chair for Dynamics, Control, Machine Learning, and Numerics (Alexander von Humboldt Professorship), Department of Mathematics,  Friedrich--Alexander-Universit\"at Erlangen--N\"urnberg,
91058 Erlangen, Germany.}

\thanks{\thankssymb{1}Departamento de Matem\'{a}ticas,
Universidad Aut\'{o}noma de Madrid,
28049 Madrid, Spain.
}

\thanks{\thankssymb{3}Chair of Computational Mathematics, Fundaci\'{o}n Deusto. Av. de las Universidades, 24,
48007 Bilbao, Basque Country, Spain.
}

\begin{document}

\begin{abstract}
Transformers are extremely successful machine learning models whose mathematical properties remain poorly understood. 
Here, we rigorously characterize the behavior of
transformers with hardmax self-attention and normalization sublayers as the number of layers tends to infinity.
By viewing such transformers as discrete-time dynamical systems describing the evolution of points in a Euclidean space, and thanks to a geometric interpretation of the self-attention mechanism based on hyperplane separation, we show that the transformer inputs asymptotically converge to a clustered equilibrium determined by special points called \textit{leaders}.  
We then leverage this theoretical understanding to solve sentiment analysis problems from language processing using a fully interpretable transformer model, which effectively captures `context' by clustering meaningless words around leader words carrying the most meaning. 
Finally, we outline remaining challenges to bridge the gap between the mathematical analysis of transformers and their real-life implementation.
\end{abstract}
\maketitle
%
%
\section{Introduction}
Transformers are fundamental machine learning models that consistently outperform other deep learning paradigms, such as recurrent neural networks~\cite{lstm1997}, in language modeling~\cite{brown2020gpt3}, computer vision~\cite{dosovitskiy2021image}, and audio processing~\cite{pmlr-v202-radford23a}.
The power of transformers is often attributed to the so-called \textit{self-attention sublayers}, which are added to the normalization and feed-forward sublayers of a standard residual neural network and which, heuristically, identify relations between components of the transformer's input data to facilitate prediction tasks. A rigorous explanation for the role of the self-attention sublayers, however, remains to be given.

\begin{figure}
    \centering
    \begin{subfigure}{.5\textwidth}
      \centering
      \includegraphics{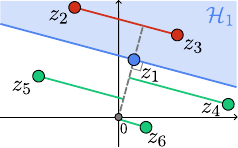}
      \caption{}
    \end{subfigure}%
    \begin{subfigure}{.5\textwidth}
      \centering
      \includegraphics{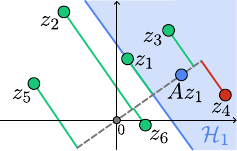}
      \caption{}
    \end{subfigure}
    \caption{Geometric interpretation of~\cref{eq:transformer_b} for $i=1$ with (a) $A=I$ and (b) $A = \left(\begin{smallmatrix} 2 & 1 \\ 1 & 1 
    \end{smallmatrix}\right)$. In (a), tokens $\zz{2}{}$ and $\zz{3}{}$ have the largest projection on the direction of $A \zz{1}{} = \zz{1}{}$, so $\CC{1}{} = \{ 2,3 \}$. In (b), token $\zz{4}{}$ has the largest projection on the direction of $A\zz{1}{}$, so $\CC{1}{} = \{ 4 \}$. In both cases, tokens attracting $z_1$ lie in the half-space $\mathcal{H}_1 = \{z:\;\langle A \zz{1}{}, \zz{}{} - \zz{1}{} \rangle \geq 0\}$ (blue shading).}
    \label{fig:geom_interp}
\end{figure}

In this work, we take a step in this direction and explain with precise mathematical results the role of self-attention for a simplified, yet expressive, class of transformers that we call \textit{pure-attention hardmax transformers}. Defined precisely in \Cref{ss:modeling}, such transformers are parameterized by a symmetric positive definite matrix $A \in \mathbb{R}^{d\times d}$ and a scalar $\alpha > 0$. They act on points $\zz{1}{},\ldots,\zz{n}{} \in \R^d$, called \textit{tokens} in the machine learning literature, that we arrange as the rows of a matrix $Z \in \mathbb{R}^{n\times d}$. In applications, tokens encode components of the transformer input, such as words in a language model or snippets of an image in computer vision. Given initial token values $\zz{1}{0},\ldots,\zz{n}{0}$, the value $\zz{i}{t+1}$ of token $\zz{i}{}$ returned by the $t$\textsuperscript{th} layer of our transformer model is
\begin{subequations}\label{eq:transformer}
\begin{equation}
\label{eq:transformer_a}
\zz{i}{t+1}=\zz{i}{t}+\frac{\alpha}{1+\alpha}\,\frac{1}{|\CC{i}{t}|}\sum_{j\in \CC{i}{t}}\Big(\zz{j}{t}-\zz{i}{t}\Big),
\end{equation}
where $Z^t$ is the token matrix in layer $t$ and
\begin{equation}
\label{eq:transformer_b}
    \CC{i}{t}= 
    \left\{ 
    j\in [n]\,\, : \,\,
    \big\langle A \zz{i}{t}, \zz{j}{t}\big\rangle
    =
    \max_{\ell\in[n]}
    \big\langle A \zz{i}{t}, \zz{\ell}{t}\big\rangle
    \right\}.
\end{equation}
\end{subequations}
Here, $|\CC{i}{t}|$ is the cardinality of the index set $\CC{i}{t}$, angled brackets denote the standard inner product in $\R^d$, and $[n]=\{1,\ldots,n\}$. Our model and its relation to other transformer models from the literature are discussed further in \Cref{sec:modelingOfTransformers}. We note only that if we view \cref{eq:transformer} as a discrete-time dynamical system describing the evolution of tokens (see \cite{lu2019understanding,geshkovski2023emergence,geshkovski_mathematical_2023} for more on this dynamical perspective), then the self-attention mechanism has a simple geometric interpretation: token $\zz{i}{}$ is attracted to the tokens with the largest orthogonal projection in the direction of $A \zz{i}{}$ (cf. \Cref{fig:geom_interp}), with $\alpha$ acting as a step-size that regulates the attraction intensity.
\subsection{Our results} Let us now summarize the main contributions of the paper.

\subsubsection{Asymptotic clustering}
\label{ss:theoretical-contribution}
Motivated by the increasing depth of transformers (GPT-3 has up to 96 layers~\cite{brown2020gpt3}), we study the asymptotic behavior of tokens that evolve according to \cref{eq:transformer} starting from initial values $\zz{1}{0}, \dots, \zz{n}{0}$. We prove that, as $t\to \infty$, tokens converge to a clustered equilibrium where the cluster points are either special tokens we call \emph{leaders}, or particular convex combinations thereof.
\begin{definition}
\label{defi:leaderIntro}
    Token $\zz{i}{}$ is a \emph{leader} if there exists $t \in \N$ such that $\CC{i}{t} = \{ i \}$.
\end{definition}
\begin{theorem}
\label{thm:emergenceClusters}
Assume the initial token values $\zz{1}{0},\ldots,\zz{n}{0} \in \R^d$ are nonzero and distinct. Assume also the matrix $A\in \R^{d\times d}$ in \cref{eq:transformer_b} is symmetric and positive definite. Then, the set of leaders $\mathcal{L}$ is not empty, and there exist a convex polytope $\mathcal{K}$ with $| \mathcal{L} |$ vertices and a finite set $\mathcal{S} \subset \partial \mathcal{K}$ such that:
\begin{enumerate}[(i)]
    \item Every token converges to a point in $\mathcal{S}$.
    \item Every leader converges to a distinct vertex of $\mathcal{K}$ in finite layers.
    \item If $s\in \mathcal{S}$ is not a vertex of $\mathcal{K}$, then it is a projection of the origin onto a face of $\mathcal{K}$ with respect to the norm associated with $A$.
\end{enumerate}
\end{theorem}
\begin{remark}
    Statement $(ii)$ implies that $\mathcal{S}$ contains the vertices of $\mathcal{K}$, which are the limiting values of the leaders and are attained in a finite number of layers.
\end{remark}
\begin{remark}
    When a token value is initially zero, \Cref{thm:emergenceClusters} still holds if one adds the origin to $\mathcal{S}$. For example, with $A = I$ and any $\alpha >0$, the $n = 5$ tokens in $\R$ with initial values $z_1^0 = -1$, $z_2=-1/2$, $z_3^0 = 0$, $z_4^0=1/2$ and $z_5^0 = 1$ converge to the cluster set $\mathcal{S} = \{ -1, 0, 1\}$. In particular, token $z_3$ remains at the origin. An initially zero token, however, remains zero only in symmetric situations that we believe are non-generic. Once an initially zero token has moved, \Cref{thm:emergenceClusters} applies as stated.
\end{remark}

\begin{figure}[t]
    \centering    
    \includegraphics[width=0.9\linewidth]{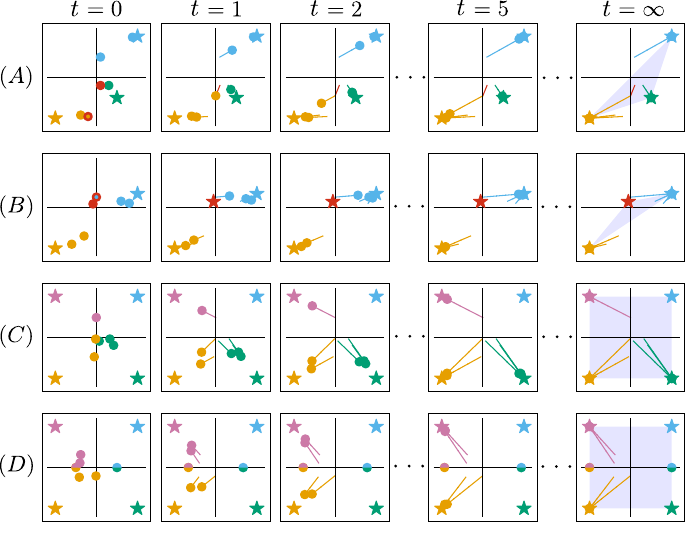}
    \caption{Simulations of \cref{eq:transformer} with $\alpha = 0.5$, $A = I$, and four different initial token values. In each panel, stars denote tokens $\zz{i}{}$ satisfying $\CC{i}{t}=\{i\}$ at layer $t\in \N$, while circles denote all other tokens. Colors indicate which tokens are being followed. Tokens painted in two halves follow two tokens. Tokens whose interior and edge colors are different, instead, follow tokens of their interior color and are followed by tokens of their edge color. The shaded region in the last column is the closed convex hull of leaders.}
    \label{fig:dependence_IC}
\end{figure}

We stress that the set of leaders $\mathcal{L}$, the set of cluster points $\mathcal{S}$, and the convex polytope $\mathcal{K}$ in \Cref{thm:emergenceClusters} depend strongly on the initial token values $\zz{1}{0},\ldots,\zz{n}{0}$ and on the transformer parameters $A$ and $\alpha$. We do not write this dependency explicitly to lighten the notation. To illustrate the dependency on the initial token values, \Cref{fig:dependence_IC} shows simulations of the model \cref{eq:transformer} for $A = I$, $\alpha = 0.5$, and four different initial token values. In row $(A)$, the cluster set consists of the three leaders, which are determined initially. The situation is similar in row $(B)$, but not all leaders are determined initially (notice the leader closest to the origin is determined at layer $t = 1$). Rows $(C)$ and $(D)$ have the same leaders and limiting convex polytope $\mathcal{K}$ (blue shading). However, in row $(C)$, the cluster set consists of the four leaders, while in row $(D)$ the cluster set also contains two extra points. These are orthogonal projections of the origin onto two faces of $\mathcal{K}$, as predicted by \Cref{thm:emergenceClusters} for $A = I$.

\Cref{thm:emergenceClusters} is proven in \Cref{sec:problemFormulation,,sec:mainResults,sec:leaders} in four steps. In the first one (\Cref{lem:shrinking,lem:Kattracting}), we show that the convex hull of the token values $\zz{1}{t},\ldots,\zz{n}{t}$ shrinks towards some convex polytope $\mathcal{K}$ as $t$ increases. (We do not know at this stage that $\mathcal{K}$ is the closed convex hull of the leaders.)
We then establish that the token values $\zz{1}{t},\ldots,\zz{n}{t}$ are attracted to a finite set of cluster points $\mathcal{S} \subset \partial\mathcal{K}$ (\Cref{lem:setSattracting}), which consists of the vertices of $\mathcal{K}$ plus projections of the origin onto the faces of $\mathcal{K}$ with respect to the norm associated to $A$ (\Cref{lem:VinS}).
In the third step of the proof, we show that tokens cannot jump indefinitely between sufficiently small neighborhoods of cluster points in $\mathcal{S}$ (\Cref{th:noCirculation}), so every token must converge to a single cluster point. In the fourth and final step, we show that the vertices of the convex polytope $\mathcal{K}$ coincide with the asymptotic values of the leaders (\Cref{lem:ownMaxImpliesVertex,,lem:vertexImpliesOwnMax}).

\subsubsection{Interpretable transformers for sentiment analysis}
\label{ss:computational-contribution}
Having shown that pure-attention hardmax transformers modeled by \cref{eq:transformer} exhibit clustering, in \Cref{sec:numerics}, we shift focus and demonstrate how this theoretical insight provides an interpretative framework for understanding the behavior of transformers in sentiment analysis. More precisely, we construct a minimal yet fully interpretable transformer model to classify movie reviews as either positive or negative. 

Our model consists of three core components: an encoder, which maps words to tokens in $\R^d$; a transformer, which updates the token values over a finite number of layers; and a decoder, which predicts either positive or negative sentiment based on the token values after the final transformer layer.
Through computational experiments, we demonstrate that the clustering behavior induced by our transformer provides context by grouping semantically insignificant words around the most meaningful ones. This allows for a clear interpretation of the role of each component in our model: the encoder identifies and selects meaningful words as leaders; the transformer clusters words around semantically meaningful leaders; and the decoder partitions the token space into two half-spaces corresponding to positive and negative sentiment, producing a prediction based on the position of the mean token value in the final layer.

We emphasize that our primary objective is to offer an interpretative framework for transformer models, rather than to enhance accuracy or training efficiency in the more complex transformer architectures commonly used in practice.

\subsection{Related work}
\label{ss:related-work}
Our contributions are part of ongoing efforts to rigorously explain the success of transformers in applications ranging from powering text-based AI platforms \cite{brown2020gpt3,openai2024gpt4technicalreport, dubey2024llama} and image generators \cite{pmlr-v139-ramesh21a, esser2021tamingTransformers, esser2024scaling} to assisting and proving mathematical theorems~\cite{polu2020generative,HayatLean, trinh2024solving}, analyzing dynamics~\cite{HayatLyapunov}, studying metabolic networks~\cite{HayatMetabolic}, and predicting protein structures~\cite{Jumper2021HighlyAP, abramson2024accurate} and chemical reactions \cite{schwaller2019molecular}.

From the perspective of approximation theory, the power of transformers lies in their ability to approximate with arbitrary accuracy any continuous equivariant sequence-to-sequence function on compact sets \cite{yunAreTransformersUniversal2020,alberti2023sumformer}.
Other authors have instead adopted the perspective of dynamics \cite{Weinan2017APO,lu2019understanding,geshkovski2023emergence,geshkovski_mathematical_2023}, viewing transformers as discretization schemes for \textit{neural ordinary differential equations} (NODEs)~\cite{chen2018}, ODEs for advecting and diffusing particles~\cite{lu2019understanding}, and the Wasserstein gradient flow equations of certain energy minimization problems~\cite{sanderSinkformers2022}. Passing to continuous-time transformer models is convenient because one can apply well-known tools for dynamical system analysis, which have already improved our understanding of deep learning models such as residual neural networks~\cite{li2022deep,geshkovski2022turnpike,domenec2023NODES}. In particular, clustering results similar to those in \Cref{thm:emergenceClusters}(i) were proven in \cite{geshkovski2023emergence,geshkovski_mathematical_2023} for NODE models of transformers without feed-forward sublayers (see \Cref{fig:transformer_diag}). There is also a similarity with asymptotic clustering opinion dynamics models, but, compared to these, transformers pose unique challenges; we refer to \cite[\S1.3.4]{geshkovski2023emergence} for further discussion.

Compared to these works, our analysis has two distinguishing features. First, we replace a `softmax' formulation of the self-attention mechanism with the `hardmax' formulation from \cref{eq:transformer_b} (see \Cref{ss:comparisonModeling} for more details). The former is usually preferred because, being a smooth regularization of the latter, it facilitates computing gradients in training algorithms and it ensures that ODE models of transformers are well-posed \cite[\S6]{geshkovski2023emergence}. On the other hand, our `hardmax' formulation has a clearer geometric interpretation that reveals the key role of leaders in the transformer dynamics. This yields a more precise description of the clustering phenomena arising in transformers compared to previous works. 
Last but not least, our `hardmax' formulation clearly prevents the emergence of metastable states, where tokens initially form clusters that merge slowly into a single point. This phenomenon is observed in `softmax' transformer models on the unit sphere of $\R^d$ (see \cite{geshkovski_mathematical_2023} for numerical evidence and \cite{geshkovski2024dynamic, bruno2024emergence} for rigorous theory). 
In contrast, our model \cref{eq:transformer}, defined on all of $\R^d$, does not exhibit metastability, as leaders stop evolving in a finite number of steps, fully determining the final clustering pattern.
Analyzing a system like ours, which stabilizes in finite time, is crucial because it allows for a genuine classification of the clustering pattern. In metastable models, this task is more complex, requiring characterization of both intermediate `metastable' configurations and their escaping times.
The second distinguishing feature is that we abandon the continuous-time framework of NODEs and, instead, work directly at the discrete level. One reason for this choice is that the connection between continuous-time transformer models and real-life transformers with discrete layers remains to be rigorously justified (see, however, the works \cite{hayou2023on,peluchetti20a,peluchetti2021} for analysis that links classical residual neural networks and stochastic versions of NODEs). Another is that, in the discrete-time setting, we avoid the well-posedness issues related to the lack of smoothness of our `hardmax' formulation for the self-attention mechanism (see \cite[\S7.2.2]{geshkovski_mathematical_2023} for a more detailed discussion). 
These advantages, however, come at the cost of technical difficulties in the proof of \Cref{thm:emergenceClusters}, which we resolve. Since our proof follows steps similar to those in~\cite{geshkovski2023emergence}, but our dynamics lack continuity of trajectories, we develop a different argument to show that tokens cannot `jump' indefinitely between elements of the attracting set $\mathcal{S}$.

Finally, we mention that the clustering analysis of hardmax transformers carried out in this work helps explain the effectiveness of transformers beyond the sentiment analysis task considered in \Cref{sec:numerics}. For example, in a follow-up work~\cite{alcalde2025exactsequence} we leverage the clustering effect to show that transformers with alternating hardmax self-attention and feed-forward layers can solve sequence classification and next-token prediction tasks exactly and using a number of parameters independent of sequence length.

\subsection{Outline}
The rest of the paper is organized as follows. In \Cref{sec:modelingOfTransformers}, we review the general modeling of transformers and derive our transformer model \cref{eq:transformer}. In \Cref{sec:problemFormulation}, we establish some basic properties of this model. In \Cref{sec:mainResults}, we prove that tokens converge to a clustered equilibrium as the number of transformer layers becomes infinite. In \Cref{sec:leaders}, we characterize the cluster points via leaders. In \Cref{sec:numerics}, we apply our clustering results to solve the real machine learning task of sentiment analysis in an interpretable way. Finally, in \Cref{sec:conclusion}, we propose a number of open problems for the mathematical analysis of transformers.

\section{Mathematical modeling of transformers}
\label{sec:modelingOfTransformers}
Transformers are complex machine learning models that are usually described through the composition of several functions with distinct roles. One might wonder, therefore, how  the model in \cref{eq:transformer} relates to general transformers deployed in applications. 
The objective of this section is to clarify this relationship. We start in \Cref{ss:modeling} by reviewing the mathematical formulation of transformers in a general setting. We then describe in \Cref{ss:pureAttention} how our transformer \cref{eq:transformer} is derived from the general model through particular modeling choices. Finally, in \Cref{ss:comparisonModeling} we briefly compare our transformer to other models analyzed in the literature.
\begin{figure}
    \centering
    \includegraphics[width = 0.9\textwidth]{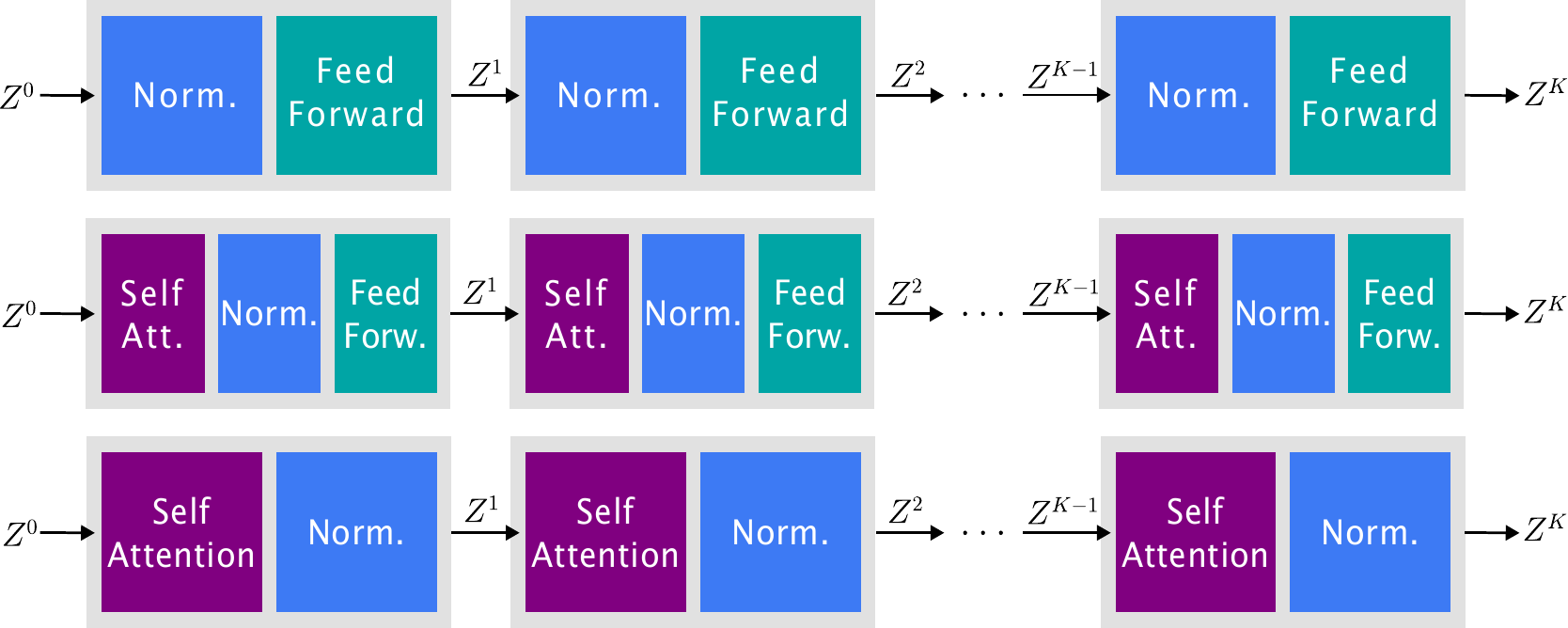}
    \caption{Schematic illustrations of a deep neural network with normalization and feed-forward sublayers (top), a full transformer with self-attention, normalization, and feed-forward sublayers (middle), and a pure-attention transformer with only self-attention and normalization sublayers (bottom). Each model takes a matrix $\ZZ^0 \in \R^{n \times d}$ as its input and outputs a matrix $\ZZ^{K} \in \R^{n \times d}$ after being processed by $K$ transformer layers. Residual connections are incorporated within each feed-forward and self-attention layer.}
    \label{fig:transformer_diag}
\end{figure}
\subsection{General mathematical models of transformers}
\label{ss:modeling}
Fix a depth $K \in \mathbb{N}$. As illustrated in the middle panel of \Cref{fig:transformer_diag}, a transformer is a machine learning model that acts on the rows $\zz{1}{0}, \dots, \zz{n}{0} \in \R^d$ of a matrix input $\ZZ^0 \in \R^{n \times d}$ via
\begin{equation}\label{e:transf-dynamics-abstract}
    \ZZ^{t+1} = \mathcal{T}_t(\ZZ^t),\qquad t \in \{0,\ldots,K-1\}.
\end{equation}
Each \textit{layer} of the transformer is a function $\mathcal{T}_t: \R^{n\times d} \to \R^{n\times d}$ defined by composing  a \textit{self-attention sublayer} $\mathcal{A}: \R^{n \times d}\to \R^{n \times d}$, a \textit{normalization sublayer} $\mathcal{N}: \R^{n \times d}\to \R^{n \times d}$, and a \textit{feed-forward sublayer} $\mathcal{F}: \R^{n \times d}\to \R^{n \times d}$, which we now describe.

The self-attention sublayer maps its input $Z \in \R^{n \times d}$ into a matrix $\mathcal{A}(Z) \in \R^{n \times d}$ whose $i$\textsuperscript{th} row is given by
\begin{equation}
\label{eq:attentionLayer}
    \mathcal{A}_i(\ZZ) = \zz{i}{} + \sum_{j=1}^n \Lambda_{ij}(\ZZ, A) V \zz{j}{}.
\end{equation}
Here, the \textit{attention} and \textit{value} matrices $A,V \in \R^{d\times d}$ are determined during training\footnote{The attention matrix is often specified as $A = Q^\top K$ for trainable \textit{query} and \textit{key} matrices $Q,K \in \R^{d \times d}$. Since the model properties depend only on the matrix $A$, we work directly with the latter. However, in training large transformers, $Q$ and $K$ are typically low-rank, which can cause the loss of definiteness assumed in $A$ (see \Cref{sec:conclusion} for further discussion).}, and $\Lambda_{ij}(Z,A)$ are the entries of an $n\times n$ row-stochastic matrix $\Lambda$, called the \textit{similarity} matrix, whose role is to capture similarities between the rows of $\ZZ$ (i.e., the tokens). Observe that, since $\Lambda_{ij}(\ZZ,A) \geq 0$ and $\smash{\sum_{j=1}^n \Lambda_{ij}(\ZZ,A) = 1}$ by assumption, the self-attention sublayer simply adds to each token $z_i$ a multiple of a weighted average of linearly transformed tokens $V z_j$.

The normalization sublayer is simply a function $\mathcal{N}: \R^{n \times d}\to \R^{n \times d}$ that rescales each token to prevent blow up and mitigate the vanishing and exploding gradient problem~\cite{pascanu2013difficulty} during the training phase.

Finally, the feed-forward sublayer transforms its input $Z\in \R^{n \times d}$ into a matrix $\mathcal{F}(Z) \in \R^{n \times d}$ whose $i$\textsuperscript{th} row is $\mathcal{F}_i(\ZZ) = z_i + U_i\, \sigma(W_i z_i + b_i) + c_i$, where $\sigma: \R^{d} \to \R^{r}$ is a user-prescribed activation function while the weight matrices $W_i \in \R^{r \times d}$, $U_i \in \R^{d\times r}$ and the bias vectors $b_i \in \R^r$, $c_i \in \R^d$ are determined through training.

\begin{remark}
    The self-attention sublayers in~\cref{eq:attentionLayer} include a `residual connection', which is standard in the modeling of transformers. This residual connection ensures that distinct tokens in the sublayer input $Z$ remain distinct in the $\mathcal{A}(Z)$ (see \Cref{lem:neverCollide}). They are also the reason why, in \Cref{thm:emergenceClusters}, clusters emerge only asymptotically as the number of layers grows. We expect our analysis to extend to self-attention sublayers without residual connections (i.e., $\mathcal{A}_i(\ZZ) = \sum_{j=1}^n \Lambda_{ij}(\ZZ, A) V \zz{j}{}$), in which case we also anticipate perfect clustering in a finite number of layers because distinct tokens can collide. 
\end{remark}

\subsection{Pure-attention hardmax transformers}
\label{ss:pureAttention} 
Having reviewed the general modeling of transformers, we now show how particular choices for the layers result in the pure-attention hardmax transformers described by \cref{eq:transformer}. 

As sketched in the bottom panel of \Cref{fig:transformer_diag}, pure-attention hardmax transformers consist of identical layers (a weight-sharing approach also present in practice \cite{Lan2020ALBERT}) with self-attention and normalization sublayers, but no feed-forward sublayers. For the attention and value matrices $A,V \in \R^{d\times d}$, we make the following assumptions.
\begin{assumption}
\label{ass:IC}
The matrices $A,V \in \R^{d\times d}$ are such that:
\begin{enumerate}[(i)]
    \item $A$ is symmetric and positive definite.
    \item $V = \alpha I$ for some real number $\alpha > 0$.
\end{enumerate}
\end{assumption}
\noindent
These two conditions are common in the mathematical analysis of transformers (see, e.g.,~\cite{geshkovski2023emergence,geshkovski_mathematical_2023, sanderSinkformers2022}). For the similarity matrix $\Lambda \in \R^{n \times n}$ in \cref{eq:attentionLayer}, we consider a `hardmax' formulation and set
\begin{equation}\label{eq:att-hardmax}
    \Lambda_{ij}(\ZZ,A) = 
    \begin{cases}
        \frac{1}{\abs{ \CC{i}{} }} &\text{if } j \in \CC{i}{},\\
        \quad 0 &\text{otherwise}
    \end{cases}
    \qquad \forall i,j \in [n],
\end{equation}
where $\CC{i}{}$ are the index sets introduced in \cref{eq:transformer_b}. For the normalization sublayer, we follow~\cite[Remark 3.4]{geshkovski2023emergence} and set
\begin{equation}
    \label{eq:normLayer-V=I}
    \mathcal{N}(\ZZ) = \frac{1}{1+\alpha} \ZZ.
\end{equation}
Thus, the normalization layer simply rescales all tokens by a constant factor depending on $\alpha$. The equations for the self-attention and normalization sublayers are easily combined to obtain what we called pure-attention hardmax transformers in \cref{eq:transformer}.

\subsection{Comparison to other simplified transformer models}
\label{ss:comparisonModeling}
To conclude this section, we briefly compare our modeling assumptions to other common choices considered in previous works~\cite{geshkovski2023emergence,geshkovski_mathematical_2023}, which result in different transformer models.

The similarity matrix $\Lambda \in \R^{n\times n}$ in  the self-attention sublayer is usually defined using a `softmax' function parametrized by a \textit{temperature} parameter $\tau>0$,
\begin{equation}\label{e:att-softmax}
    \Lambda_{ij}(\ZZ, A) = 
    \dfrac{
        {\rm e}^{\frac1\tau\left\langle A \zz{i}{}, \zz{j}{} \right\rangle}
    }{
        \displaystyle\sum_{k=1}^n {\rm e}^{\frac1\tau \left\langle A \zz{i}{}, \zz{k}{} \right\rangle}
    }
    \qquad \forall i,j \in [n].
\end{equation}
This formulation has the benefit of being smooth with respect to the matrix $A$, which facilitates the training of the model. When the tokens are assumed to be in a compact set, then the hardmax formulation is formally obtained by letting the temperature parameter $\tau\to 0$~\cite{elfadel1993softmax}. In this sense, the softmax formulation is understood as a regularization of the hardmax formulation.

For the normalization sublayer, an alternative to \cref{eq:normLayer-V=I} is to project tokens onto the unit sphere of $\R^d$ \cite{batchNormalization2015,geshkovski_mathematical_2023}. For our hardmax formulation, such projection is of little interest, as one can check using the arguments in \Cref{lem:BasicPropertiesLeaders}(i) that if all tokens have initially the same norm, then they remain fixed for all times. Therefore, we choose the mathematically simpler and dynamically more interesting form \cref{eq:normLayer-V=I}.
\section{Preliminary observations} 
\label{sec:problemFormulation}
This section establishes some basic properties and preliminary results of our model \cref{eq:transformer}, which will be used extensively in the proofs of \Cref{sec:mainResults,sec:leaders}. Throughout the paper, we assume that the initial token values are distinct and nonzero, as assumed in \Cref{thm:emergenceClusters}. To simplify the notation and the geometric intuition behind our analysis, we will also fix $A = I$. However, that \Cref{thm:emergenceClusters} holds for general symmetric positive definite $A$ can be seen from a change of variables. Indeed, factorizing $A = B^\top B$ with $B \in \R^{d\times d}$ invertible, one finds that the transformed tokens $\Tilde{z}_i = B \zz{i}{}$ evolve according to \cref{eq:transformer} with $\mathcal{C}_i(\Tilde{Z})$ defined using the identity matrix in place of $A$. If, as we prove, the transformed tokens cluster, then so do the original ones $\zz{1}{} = B^{-1} \Tilde{z}_1, \dots, \zz{n}{} = B^{-1} \Tilde{z}_n$. Finally, in keeping with the dynamical systems interpretation of transformers \cite{geshkovski2023emergence,geshkovski_mathematical_2023}, we view the layer index $t$ as a time variable and refer to $\zz{i}{t}$ as the value of token $\zz{i}{}$ at time $t$.
\subsection{Tokens do not collide}
We start by showing that if the initial token values are distinct, then they remain distinct for all times.
\begin{lemma}
\label{lem:neverCollide}
If the initial token values $ z_1^0, \dots, z_n^0 \in \R^{d}$ are distinct, then $\zz{i}{t} \neq \zz{j}{t}$ for all $i\neq j$ and for all $t\in \N$.
\end{lemma}
\begin{proof}
It suffices to prove that $z_i^1 \neq z_j^1$ for all $i\neq j$, since the argument can be iterated. Arguing by contradiction, suppose that there exist tokens $z_i$ and $z_j$ with distinct initial values such that $z_i^1 = z_j^1$. Using the model \cref{eq:transformer} to express $z_i^1$ and $z_j^1$ as a function of the initial values $z_1^0, \dots, z_n^0$, we find that
\begin{equation}
    \frac{1}{1+\alpha}\zz{i}{0}+
    \frac{\alpha}{1+\alpha}\,\frac{1}{|\CC{i}{0}|}
    \sum_{\ell\in \CC{i}{0}}\zz{\ell}{0}
    = 
    \frac{1}{1+\alpha}\zz{j}{0}+
    \frac{\alpha}{1+\alpha}\,\frac{1}{|\CC{j}{0}|}\sum_{r\in \CC{j}{0}}\zz{r}{0}.
\end{equation}
Upon reordering the terms, multiplying by $1 + \alpha$, and taking the inner product of both sides with $z_i^0 - z_j^0$, we obtain
\begin{align}
    \label{eq:collide1}
    \left\| \zz{i}{0} -\zz{j}{0} \right\|^2 
    &= 
    -\alpha \Bigg< \frac{1}{|\CC{i}{0}|}\sum_{\ell\in \CC{i}{0}}\zz{\ell}{0} - \frac{1}{|\CC{j}{0}|}\sum_{r\in \CC{j}{0}}\zz{r}{0} , \zz{i}{0} - \zz{j}{0} \Bigg> \\
    &= - \frac{\alpha}{|\CC{i}{0}||\CC{j}{0}|} \sum_{\ell\in \CC{i}{0}} \sum_{r\in \CC{j}{0}} \left< \zz{\ell}{0} - \zz{r}{0} , \zz{i}{0} - \zz{j}{0} \right>. \nonumber
\end{align}
We now observe that $\langle \zz{\ell}{0} - \zz{r}{0} , \zz{i}{0}\rangle \geq 0$ because $\ell \in \CC{i}{0}$ and, similarly, $\langle \zz{r}{0} - \zz{\ell}{0} , \zz{j}{0}\rangle \geq 0$ because $r \in \CC{j}{0}$. We thus have
$\langle \zz{\ell}{0} - \zz{r}{0} , \zz{i}{0} - \zz{j}{0} \rangle = \langle \zz{\ell}{0} - \zz{r}{0}  , \zz{i}{0}\rangle + \langle \zz{r}{0} - \zz{\ell}{0} , \zz{j}{0}\rangle \geq 0$ and conclude that the right-hand side of \cref{eq:collide1} is non-positive. This is possible only if $\| \zz{i}{0} - \zz{j}{0} \| = 0$, which contradicts the assumption that $\zz{i}{0} \neq \zz{j}{0}$.
\end{proof}

\subsection{The norm of tokens cannot decrease}
We next prove that the norm of tokens cannot decrease in time. This property, combined with \Cref{lem:neverCollide}, implies that tokens with initial values distinct and nonzero, as assumed in \Cref{thm:emergenceClusters}, remain distinct and nonzero for all future times.
\begin{lemma}
\label{lem:normIsGrowing}
For every token $z_i$ and time $t \in \N$, $\| \zz{i}{\ell} \| \geq \|\zz{i}{t} \|$ for all $\ell \geq t$.
\end{lemma}
\begin{proof}
It suffices to fix $t = 0$ and $\ell = 1$, and show $\| \zz{i}{1} \|^2 - \| \zz{i}{0} \|^2 \geq 0$. Using the model \cref{eq:transformer}, and setting $\nu_i^0 \coloneqq \alpha (1 + \alpha)^{-1}|\CC{i}{0}|^{-1}$ to lighten the notation, we have
\begin{align}
    \label{eq:calculationsNorm1}
        \big\| \zz{i}{1} \big\|^2 - \big\| \zz{i}{0} \big\|^2 &= \langle \zz{i}{1} - \zz{i}{0}, \zz{i}{1} +  \zz{i}{0}\rangle 
        \\
        &= \bigg<
        \nu_i^0 \sum_{j\in \CC{i}{0}} (\zz{j}{0}-\zz{i}{0}), \; 2\zz{i}{0} +  \nu_i^0 \sum_{j\in \CC{i}{0}}(\zz{j}{0}-\zz{i}{0}) \bigg>
        \nonumber \\
        &= 2\nu_i^0 \sum_{j\in \CC{i}{0}}\langle \zz{j}{0}-\zz{i}{0} , \zz{i}{0} \rangle + \bigg\| \nu_i^0 \sum_{j\in \CC{i}{0}} (\zz{j}{0}-\zz{i}{0}) \bigg\|^2.\nonumber
    \end{align}
The last expression is non-negative since $\langle \zz{j}{0}-\zz{i}{0} , \zz{i}{0} \rangle \geq 0$ by definition of $\CC{i}{0}$. 
\end{proof}
\subsection{Shrinking of the convex hull of tokens}
The normalization sublayer \cref{eq:normLayer-V=I} is particularly convenient because, as we show next, the convex hull of the token values $\zz{1}{t},\ldots,\zz{n}{t}$ does not increase in time in the sense of set inclusion. In particular, tokens remain in the (compact) closed convex hull of the initial token values at all times. We will denote by $\overline{\text{co}}(\ZZ^{t})=\left\{ z=\sum_{i=1}^{n}\lambda_{i}\zz{i}{t}\quad\text{where}\quad\sum_{i=1}^{n}\lambda_{i}=1,\;\;\lambda_{i}\geq0\right\}$ the closed convex hull of the token values $\zz{1}{t},\ldots,\zz{n}{t}$ at time $t$.
\begin{lemma}
\label{lem:shrinking}
For every time $t\in \N$, $\co (\ZZ^{t + 1}) \subseteq \co (\ZZ^{t})$.
\end{lemma}
\begin{proof}
For every token value $\zz{i}{t+1}$, our model \cref{eq:transformer} is equivalently written as
\begin{equation}
    \zz{i}{t+1}=\frac{1}{1 + \alpha}\zz{i}{t}+\frac{\alpha}{1+\alpha}\,\frac{1}{|\CC{i}{t}|}\sum_{j\in \CC{i}{t}}\zz{j}{t}.
\end{equation}
It is straightforward to verify that $\zz{i}{t+1}$ is a convex combination of $\zz{1}{t},\ldots,\zz{n}{t}$. This implies $\zz{i}{t+1} \in \co (\ZZ^t)$ and proves $\co (\ZZ^{t+1}) \subseteq \co (\ZZ^{t})$, as desired.
\end{proof}
As a direct consequence \Cref{lem:shrinking}, we obtain that tokens are bounded for all times.
\begin{corollary}
\label{cor:boundedTokens}
Let $\overline{B_R}$ be the closed ball centered at the origin of radius $R = \max_{\ell \in [n]} \| \zz{\ell}{0} \|$. Then, for every token $\zz{i}{}$ and time $t \in \N$, $\zz{i}{t} \in \co (\ZZ^0)$ and $\zz{i}{t} \in \overline{B_R}$.
\end{corollary}
\begin{proof}
     By \Cref{lem:shrinking}, $\zz{i}{t} \in \co (\ZZ^t) \subseteq \co(\ZZ^0)$. Now, $\co (\ZZ^0)$ is the smallest convex set containing the initial token values and $\zz{i}{0} \in \overline{B_R}$, so $\co (Z^0) \subset \overline{B_R}$.
\end{proof}
A second important consequence of \Cref{lem:shrinking} is that the set $\co (\ZZ^t)$ converges in time to a \textit{convex polytope}, i.e., a bounded subset of $\R^d$ resulting from the intersection of finitely many half-spaces. The proof is rather elementary \cite{borovikov_intersection_1952}, but we included for completeness and because its ideas will be useful later (see the proof of \Cref{lem:VinS}).
\begin{lemma}
\label{lem:Kattracting}
The set $\mathcal{K} = \bigcap_{\ell \in \N} \co (\ZZ^\ell)$ is a non-empty convex polytope.
\end{lemma}
\begin{proof}
Since $\mathcal{K}$ is the intersection of closed, bounded, convex, and nested subsets of $\R^d$, then it is a non-empty convex set. There remains to show that $\mathcal{K}$ is a polytope. For this, notice that $\| Z^t\|_F \leq \sqrt{n} R$ where $\| \cdot\|_F$ denotes the Frobenius norm and $R$ is as in \Cref{cor:boundedTokens}. Since $Z^t$ remains in a compact subset of $\R^{n \times d}$, there exists a subsequence $\{\ZZ^t\}_{t\in I}$ converging to some limit $\ZZ \in \R^{n \times d}$. Thus, we obtain $\co (\ZZ) = \bigcap_{\ell \in I}\co (\ZZ^\ell) = \bigcap_{\ell \in \N}\co (\ZZ^\ell) = \mathcal{K}$, using that $\{\co (\ZZ^t)\}_{t\in \N}$ is nested.
\end{proof}
In the sequel, $\mathcal{K}$ will denote the limiting convex polytope $\bigcap_{\ell \in \N} \co (\ZZ^\ell)$. Even though this object depends on the initial token values, we simplify the notation by not writing such dependence explicitly. As usual, $\partial \mathcal{K}$ will denote its boundary.
\section{Emergence of clusters}
\label{sec:mainResults}
We now turn to study the asymptotic behavior of tokens, providing the necessary tools to prove convergence of all tokens to a clustered equilibrium (\Cref{thm:emergenceClusters}(i)). We begin in \Cref{subsec:attractingSet} by introducing a candidate attracting set $\mathcal{S}$. We show that this set is a finite subset of the boundary of $\mathcal{K}$. We then prove that the norm of tokens must grow strictly when not close to $\mathcal{S}$. This, combined with the boundedness of all tokens (\Cref{cor:boundedTokens}), implies that $\mathcal{S}$ is indeed attracting for our model \cref{eq:transformer}. In \Cref{subsec:particlesCluster}, we show that tokens cannot circulate indefinitely around the attracting set, but must instead converge to an equilibrium.
\subsection{The attracting set}
\label{subsec:attractingSet}
\begin{figure}[t]
    \centering
    \includegraphics[scale = 0.85]{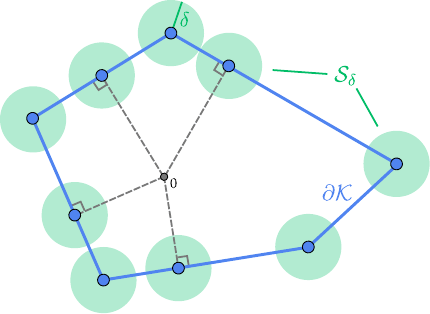}
    \caption{Sketch of the $\delta$-neighborhood of the attracting set $\mathcal{S}$.}
    \label{fig:S_delta}
\end{figure}

Let $\mathcal{K}$ be the convex polytope from \Cref{lem:Kattracting} and $\mathcal{V} = \{ v_1, \dots, v_m\}$ the set of its $m\leq n$ vertices. We define our candidate set as
\begin{equation}
\label{eq:S}
    \mathcal{S} =  \left\{ x\in \mathcal{K} : \,\, \max_{\ell \in [m]} \left< v_\ell - x, x \right> = 0 \right\}.
\end{equation}
The next result shows that $\mathcal{S}$ is a finite set on the boundary of $\mathcal{K}$.
\begin{lemma}
\label{lem:characterisationS}
The set $\mathcal{S}$ in \cref{eq:S} is a finite subset of $\partial \mathcal{K}$.
\end{lemma}
\begin{proof} 
To prove that $\mathcal{S}\subset \partial \mathcal{K}$, fix $x\in \mathcal{S}$. By definition, $\langle x,x\rangle = \max_{ \ell \in [m]} \langle v_\ell, x \rangle = \max_{ z \in \mathcal{K}} \langle z, x \rangle$, so $x$ maximizes $z \mapsto \langle z,x\rangle$ over $\mathcal{K}$. This implies $x\in \partial \mathcal{K}$, as the maximum of a linear function over a convex set is attained at the boundary.

Next, we prove that $\mathcal{S}$ is finite. By definition, every element $x$ in $\mathcal{S}$ can be written as $x = \sum_{j \in \mathcal{J}} \beta_j v_j$ where $\sum_{j\in\mathcal{J}} \beta_j = 1$ and $\beta_j \in (0,1]$ for some index set $\mathcal{J}\subset [m]$. Our objective now is to show that $x$ is the unique element in $\mathcal{S}$ which is a convex combination of $\{v_j\}_{j\in\mathcal{J}}$, for this fixed index set $\mathcal{J}$. Arguing by contradiction, suppose there exists some $y\in \mathcal{S}$, $y\neq x$ such that $y = \sum_{j \in \mathcal{J}} \gamma_j v_j$, where $\sum_{j\in \mathcal{J}} \gamma_j = 1$ and $\gamma_j \in (0,1]$. For any $i\in \mathcal{J}$, $\langle v_i, y \rangle \leq \max_{\ell \in [m]} \langle v_\ell, y\rangle = \langle y,y\rangle$ because $y\in \mathcal{S}$. Moreover, if $\langle v_i, y \rangle < \langle y, y \rangle$, then
\begin{align}
    \langle y, y \rangle = \Big< \sum_{j \in \mathcal{J}} \gamma_j v_j, y \Big> = \sum_{j \in \mathcal{J}} \gamma_j \langle v_j, y \rangle < \sum_{j \in \mathcal{J}} \gamma_j \langle y, y \rangle =  \langle y, y \rangle,
\end{align}
which is a contradiction. This yields $\langle v_i, y \rangle = \langle y, y \rangle$ for all $i\in\mathcal{J}$.
Following exactly the same argument with $y$ instead of $x$, we obtain $\langle v_i, x \rangle = \langle x, x \rangle$ for all $i \in \mathcal{J}$. Then,
\begin{align}
    \langle x, y \rangle = \Big< \sum_{j \in \mathcal{J}} \beta_j v_j, y \Big> = \sum_{j \in \mathcal{J}} \beta_j \langle v_j, y \rangle = \langle y,y\rangle, \\
    \langle x, y \rangle = \Big< x, \sum_{j \in \mathcal{J}} \gamma_j v_j \Big> = \sum_{j \in \mathcal{J}} \gamma_j \langle x, v_j \rangle = \langle x,x\rangle. 
\end{align}

In conclusion, we find that $\langle x,x\rangle = \langle x,y\rangle = \langle y,y\rangle$. This implies $\| x - y \|^2 = \langle x,x\rangle - 2 \langle x,y\rangle + \langle y,y\rangle = 0$, so $x = y$, which is a contradiction.

We end the argument by observing that, as $\mathcal{J} \subset [m]$, there are at most $2^m -1$ possible non-trivial choices for the index set $\mathcal{J}$. Since for every choice of $\mathcal{J}$ there is at most one element in $\mathcal{S}$, there is a finite number of elements in $\mathcal{S}$.
\end{proof}
Recall from \Cref{lem:normIsGrowing} that the norm of tokens does not decrease. The key property of $\mathcal{S}$ is that if a token is not close to $\mathcal{S}$, then its norm \textit{strictly} increases. To make this idea precise, for any $\delta > 0$ let us define the $\delta$-neighborhood of $\mathcal{S}$ as $\mathcal{S}_\delta = \left\{ x\in \R^d : \,\, \dist(x,\mathcal{S}) \leq \delta \right\}$, where $\dist (x,\mathcal{S}) = \min_{s\in \mathcal{S}} \| x-s\|$ denotes the distance of $x$ from $\mathcal{S}$ (cf. \Cref{fig:S_delta}).
\begin{lemma}
\label{lem:normStrictlyGrowing}
For every $\delta > 0$, there exist a time $t_\delta \in \N$ and $c>0$ depending only on $\delta$, $\mathcal{K}$ and $\alpha$ such that if $t \geq t_\delta$ and $\zz{i}{t} \notin$ $\mathcal{S}_\delta$, then $\| \zz{i}{t+1} \|^2 \geq \| \zz{i}{t}\|^2 + c$.
\end{lemma}
\begin{proof}
Define the function $f: \R^d \to \R$ as $f(x) = \max_{\ell \in [m]} \langle v_{\ell} - x, x \rangle$. Now,
\begin{equation}
    \frac{1}{|\CC{i}{t}|}\sum_{j\in \CC{i}{t}}\langle \zz{j}{t}-\zz{i}{t} , \zz{i}{t} \rangle = \max_{\ell \in [n]} \langle \zz{\ell}{t}-\zz{i}{t} , \zz{i}{t} \rangle,
\end{equation}
by definition of $\CC{i}{t}$. Combining this with \cref{eq:calculationsNorm1} yields
\begin{multline}
\label{eq:calcNorm}
    \big\| \zz{i}{t+1} \big\|^2 - \big\| \zz{i}{t} \big\|^2 
    \geq \frac{2\alpha}{1 + \alpha }\max_{\ell \in [n]} \langle \zz{\ell}{t}-\zz{i}{t} , \zz{i}{t} \rangle 
    \\         
    = \frac{2\alpha}{1 + \alpha} \max_{z \in \co (\ZZ^t)} \langle z-\zz{i}{t} , \zz{i}{t} \rangle
    \geq \frac{2\alpha}{1 + \alpha }\max_{z \in \mathcal{K}} \langle z - \zz{i}{t}, \zz{i}{t} \rangle
    =\frac{2\alpha}{1 + \alpha }f(\zz{i}{t}),
\end{multline}
where the second inequality holds because $\mathcal{K} \subset \co(\ZZ^t)$, and the last equality because the maximum over $\mathcal{K}$ is equal to the maximum over its extreme points $v_1, \dots, v_m$. Next, observe that $f$ is continuous. It is also positive on the compact set $\mathcal{K} \setminus\mathcal{S}_\delta$ by the definition of $\mathcal{S}$. Thus, we obtain $f(x) \geq c'>0$ for $x\in\mathcal{K} \setminus\mathcal{S}_\delta$. Now, set $\mathcal{K}_\varepsilon = \left\{ x\in \R^d : \,\, \dist(x,\mathcal{K}) \leq \varepsilon \right\}$. By continuity, there exists $\varepsilon_\delta > 0$ such that $f(x) \geq {c'}/{2}$ for every $x\in \mathcal{K}_{\varepsilon_\delta} \setminus \mathcal{S}_\delta$. Moreover, since $z_i^t \rightarrow \mathcal{K}$ as $t\rightarrow \infty$ by \Cref{lem:Kattracting}, there exists a time $t_\delta \in \N$ such that $\zz{i}{t} \in \mathcal{K}_{\varepsilon_\delta}$ for all $t\geq t_\delta$ and all $i\in [n]$. If $\zz{i}{t} \notin \mathcal{S}_\delta$, we obtain $f(\zz{i}{t}) \geq {c'}/{2}$, which combined with \cref{eq:calcNorm}, gives the desired inequality with $c = \alpha c'/(1+\alpha) > 0$.
\end{proof}
We are ready to show that all tokens converge to $\mathcal{S}$.
\begin{prop}
\label{lem:setSattracting}
For every token $z_i$, $\zz{i}{t} \to \mathcal{S}$ as $t\rightarrow \infty$.
\end{prop}
\begin{proof}
We need to show that, for arbitrary $\delta > 0$, there exists a time $t^\star \in \N$ such that $\zz{i}{t} \in \mathcal{S}_\delta$ for all $t\geq t^\star$. Arguing by contradiction, assume there exists an increasing and unbounded sequence of times $\{t_j\}_{j\in\N}$ such that $\zz{i}{t_j}\notin \mathcal{S}_\delta$ for all $j\in \N$. Without loss of generality, we can assume $t_j \geq t_\delta$, where $t_\delta$ is determined as in \Cref{lem:normStrictlyGrowing}. We then conclude using the same lemma that $\| z_i^{t_j + 1} \|^2 \geq\|z_i^{t_j}\|^2 + c$ for all $j\in \N$. Moreover, by \Cref{lem:normIsGrowing}, we also have $\| z_i^{t_j} \|^2 \geq\|z_i^{t_{j-1} + 1}\|^2$. Combining the last two inequalities, we obtain
\begin{equation}
    \big\| z_i^{t_j + 1} \big\|^2 \geq\big\|z_i^{t_j}\big\|^2 + c \geq \big\|z_i^{t_{j-1} + 1}\big\|^2 + c \geq
    \dots \geq \big\|z_i^{t_0}\big\|^2 + (j + 1)c.
\end{equation}
We also know from \Cref{cor:boundedTokens} that $\| \zz{i}{t}\| \leq R$ for all $t\in \N$, so $R^2 \geq \|z_i^{t_0}\|^2 + (j+1)c$. Since everything is fixed but $j$, we obtain a contradiction as $j\rightarrow\infty$.
\end{proof}
Establishing that $\mathcal{S}$ is attracting allows us to characterize this set further.
\begin{lemma}
\label{lem:VinS}
The vertices of $\mathcal{K}$ are included in the cluster set $\mathcal{S}$. Moreover, every point in $\mathcal{S}$ that is not a vertex, is a projection of the origin onto a face of $\mathcal{K}$.
\end{lemma}
\begin{proof}
To prove the first statement, we argue by contradiction and suppose there is some vertex $v\in \mathcal{V}$ such that $v \notin \mathcal{S}$. In this case, $\dist (v, \mathcal{S}) > 0$. As in the proof of \Cref{lem:Kattracting}, since $v$ is a vertex, there exists some token $z_i$, $i\in [n]$, and an increasing and unbounded sequence of times $\{t_j\}_{j\in\N}$ such that $\| v - z_i^{t_j}\| \rightarrow 0$ as $j\rightarrow \infty$. On the other hand, by \Cref{lem:setSattracting} we know that $\zz{i}{t} \rightarrow \mathcal{S}$ as $t \rightarrow \infty$, so in particular $\dist (z_i^{t_j}, \mathcal{S}) \rightarrow 0$ as $j\rightarrow \infty$. Using the triangle inequality, we obtain
    \begin{equation}
        0 < \dist (v, \mathcal{S}) \leq \big\| v - z_i^{t_j} \big\| + \dist(z_i^{t_j},\mathcal{S}) \rightarrow 0 \; \; \text{as} \;\; j\rightarrow \infty,
    \end{equation}
which is a contradiction.

We now turn to the proof of the second statement. Fix $s\in \mathcal{S} \setminus \mathcal{V}$. From \Cref{lem:characterisationS} we know that $s \in \partial \mathcal{K}$, and since $\mathcal{K}$ is a convex polytope with a finite number of faces, denote by $\{ v_j \}_{j\in \mathcal{J}} \subset \mathcal{V}$ the vertices generating a face containing $s$. We can then write
\begin{equation}
    s = \sum_{j \in \mathcal{J}} \beta_j v_j, \quad \text{where} \quad \sum_{j\in \mathcal{J}} \beta_j = 1, \quad \beta_j \in (0,1).
\end{equation}
Now, by definition of $s\in\mathcal{S}$, we have $\max_{\ell \in [m]} \langle v_\ell, s \rangle = \| s \|^2$. Denote by $\mathcal{I}$ the set of $r\leq m$ vertices indices where the maximum $\| s \|^2$ is attained. We claim that $i\in \mathcal{J}$ implies $i \in \mathcal{I}$. Indeed, if $i \notin \mathcal{I}$, then $\langle v_i, s \rangle < \| s \|^2$ and we obtain the contradiction
\begin{equation}
    \| s \|^2
    = 
    \sum_{j\in \mathcal{J}} \beta_j \langle v_j,s \rangle
    <
    \sum_{j\in \mathcal{J}} \beta_j \langle s,s \rangle 
    =
    \| s \|^2.
\end{equation}
Therefore, we have the representation $s = \sum_{j \in \mathcal{I}} \beta_j v_j$, where $\sum_{j \in \mathcal{I}} \beta_j = 1$, $\beta_j \in (0,1)$. Let $M \in \R^{r \times r}$ be the matrix with entries $M_{ij} = \langle v_i, v_j \rangle$, and $\beta = (\beta_1, \dots, \beta_r)^T \in \R^r$. Then, the pair $(x, \lambda) = (\beta, - \| s \|^2)$ solves 
the linear equations $ M x + \mathbf{1} \lambda=0$ and $\mathbf{1}^\top x = 1$, where $\mathbf{1} \in \R^r$ is the vector whose entries are equal to one. These equations are the optimality conditions of the minimization problem
\begin{equation}
    \min_{x} \big\| \sum_{j = 1}^r x_j v_j \big\|^2 \quad \text{such that} \quad \sum_{j= 1}^r x_j = 1,
\end{equation}
which yields the unique projection of the origin onto the affine space generated by $\{v_i\}_{i\in \mathcal{I}}$. The point $s$ must therefore coincide with this projection.
\end{proof}
\subsection{Tokens cluster}
\label{subsec:particlesCluster}
\begin{figure}[t]
    \centering
    \includegraphics[scale = 0.9]{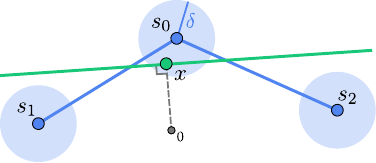}
    \caption{Geometric intuition behind \Cref{lem:verticesSelfMax} for $\mathcal{S'} = \{ s_1, s_2\}$. For $\delta > 0$  small enough and all $x\in B_\delta (s_0)$, the hyperplane with normal direction $x$ passing through $x$ (green) separates $s_0$ from the balls $B_\delta (s_1)$ and $B_\delta (s_2)$.}
    \label{fig:sketchVSelfMax}
\end{figure}

\Cref{lem:setSattracting} shows that the set $\mathcal{S}$ is attracting for our model \cref{eq:transformer}. Next, we show that tokens cannot circulate around the attracting set $\mathcal{S}$ indefinitely, and thus our model \cref{eq:transformer} converges to an equilibrium. The proof relies strongly on the following auxiliary lemma, whose meaning is illustrated in \Cref{fig:sketchVSelfMax}.
\begin{lemma}
\label{lem:verticesSelfMax}
    Fix $s_0 \in \mathcal{S}$ and $ \mathcal{S}' \subseteq \mathcal{S}\setminus \{s_0\}$. If $\langle s, s_0 \rangle  < \langle s_0, s_0 \rangle$ for all $s\in \mathcal{S}'$, there exists $\delta > 0$ such that
    $\langle y, x \rangle < \langle x, x \rangle$     
    for all $x\in B_\delta (s_0)$ and $y \in \bigcup_{s\in \mathcal{S'}} B_\delta (s)$.
\end{lemma}
\begin{proof}
Fix $\delta > 0$ small enough that $\big( 2\max_{s^\star\in \mathcal{S}} \|s^\star \| + \delta \big)\delta < {\varepsilon}/{2}$. Fix $y \in B_\delta(s)$ for $s\in \mathcal{S}'$. By hypothesis, there exists $\varepsilon > 0$ such that $\langle s, s_0 \rangle \leq \langle s_0, s_0\rangle - \varepsilon$. Then,
\begin{align}
\label{eq:ownmax1}
    \langle y, x \rangle &= \langle y - s, x - s_0 \rangle + \langle y - s, s_0 \rangle  + 
    \langle s, x - s_0\rangle +
    \langle s, s_0 \rangle \\
    &\leq \underbrace{\| y - s \| \| x - s_0\|}_{\leq \delta^2}  + \underbrace{\| y - s \|}_{\leq \delta} \| s_0\|  + \| s\| \underbrace{\| x - s_0 \|}_{\leq \delta} + \| s_0\|^2 - \varepsilon \nonumber \\
    &\leq \Big(\delta + \| s_0\| 
    + \| s \| \Big)\delta
    + \| s_0\|^2  - \varepsilon \nonumber \\
    &\leq \Big( 2\max_{s^\star \in \mathcal{S}} \|s^\star\| +\delta \Big)\delta  + \| s_0\|^2  -\varepsilon. \nonumber
\end{align}
On the other hand,
\begin{align}
\label{eq:ownmax2}
    \langle x, x \rangle 
    &= \| s_0 \|^2 + \langle x + s_0, x - s_0 \rangle  \\
    &\geq \| s_0 \|^2 - \| (x - s_0) + 2s_0 \| \| x - s_0\| \nonumber \\
    &\geq \| s_0 \|^2 - \Big( \delta + 2 \| s_0 \| \Big)\delta  \nonumber \\
    &\geq \| s_0\|^2 - \Big( 2\max_{s^\star\in \mathcal{S}} \|s^\star \| + \delta \Big)\delta. \nonumber
\end{align}
Then, \cref{eq:ownmax1} implies $\langle y, x \rangle < \| s_0\|^2 -{\varepsilon}/{2}$ while \cref{eq:ownmax2} implies $\langle x, x \rangle > \| s_0\|^2 -{\varepsilon}/{2}$. Combining both inequalities we obtain $\langle y, x \rangle < \langle x, x \rangle$, as desired.
\end{proof}

\begin{figure}[t]
    \centering
    \includegraphics[scale = 0.9]{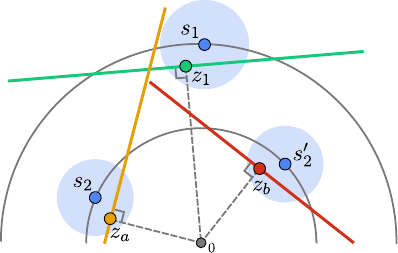}
    \caption{Sketch of the construction used to prove \Cref{th:noCirculation}. Token $z_i$ remains in a neighborhood of $s_1\in \mathcal{S}_1$. Token $z_a$ falls in case (a) of the analysis, so it remains in a neighborhood of $s_2\in \mathcal{S}_2$, while token $z_b$ falls in case (b), jumping from a neighborhood of $s_2'\in \mathcal{S}_2$ to the neighborhood of $s_1$, where it remains for all future times.}
    \label{fig:sketchConvergenceSi}
\end{figure}

We next prove that tokens must remain in a ball that contains a single element of the attracting set $\mathcal{S}$. This, together with the convergence of tokens to $\mathcal{S}$ established in \Cref{lem:setSattracting}, implies the desired clustering result.
\begin{prop}
\label{th:noCirculation}
There exists $\delta>0$ such that the following two conditions hold:
    \begin{enumerate}[(i)]
    \item If $s, s' \in \mathcal{S}$ and $s \neq s'$, then $B_\delta(s) \cap B_\delta(s') = \emptyset$.
    \item For all $i \in [n]$, there exists $s_i \in \mathcal{S}$ and $t_i \in \mathbb{N}$ such that $\zz{i}{t} \in B_\delta(s_i)$ $\forall t\geq t_i$.
    \end{enumerate}
\end{prop}
\begin{proof}
Since $\mathcal{S}$ is finite by \Cref{lem:characterisationS}, there exists $\delta_0 > 0$ such that $(i)$ is satisfied for all $\delta \leq \delta_0$. We now find $\delta \leq \delta_0$ such that $(ii)$ holds following a constructive argument. Denote by $\mathcal{S}_1 = \left\{ s_1 \in \mathcal{S}\, : \, \|s_1 \| = \max_{s \in \mathcal{S}} \| s \| \right\}$ the set of points in $\mathcal{S}$ with maximum norm. Fix $s_1\in \mathcal{S}_1$. For all $s\in \mathcal{S}\setminus \{ s_1 \}$, there holds
\begin{equation}
    \label{eq:aux_proof_conv}
    \langle s, s_1\rangle = \| s_1 \|^2 -\| s_1 \|^2 + \langle s, s_1 \rangle \leq \| s_1 \|^2 - \frac{1}{2}\| s_1 - s\|^2 < \langle s_1, s_1\rangle,
\end{equation}
where in the first inequality we have used that $\| s\| \leq \| s_1 \|$ by definition of $\mathcal{S}_1$. We can thus apply \Cref{lem:verticesSelfMax} with $s_0 = s_1$ and $\mathcal{S}' = \mathcal{S}\setminus \{ s_1 \}$ to find $\delta_1 \in (0, \delta_0]$ such that
\begin{equation}
    \label{eq:ownmax0} 
    \langle y, x \rangle < \langle x, x \rangle
\end{equation}
for all $x\in B_{\delta_1} (s_1)$ and $y \in \bigcup_{s\in \mathcal{S}'} B_{\delta_1} (s)$. By \Cref{lem:setSattracting}, there exists a time $t_1\in \N$ such that $\zz{i}{t} \in \mathcal{S}_{\delta_1}$ for all $t \geq t_1$ and all $i\in [n]$. Then, combining \cref{eq:ownmax0} with the model \cref{eq:transformer} implies that a token $\zz{i}{t} \in B_{\delta_1}(s_1)$ can only be attracted to tokens $\zz{j}{t}\in B_{\delta_1}(s_1)$, and so $z_i^t \in B_{\delta_1}(s_1)$ for all $t\geq t_1$. This argument can be applied to every point in $\mathcal{S}_1$, obtaining a collection of radii $\delta_1^1, \dots, \delta_1^{|\mathcal{S}_1|}$ and times $t_1^1, \dots, t_1^{|\mathcal{S}_1|}$. We abuse notation and re-use $\delta_1$ to denote the smallest of all such radii and $t_1$ to denote the maximum of all such times.

Next, let $\mathcal{S}_2 \coloneqq \left\{ s_2 \in \mathcal{S} \setminus \mathcal{S}_1 \,: \, \|s_2\| = \max_{s \in \mathcal{S}\setminus \mathcal{S}_1 } \| s \| \right\}$ be the set of points in $\mathcal{S}$ but not in $\mathcal{S}_1$ with maximum norm. Fix $s_2\in \mathcal{S}_2$. Then, for all $s\in \mathcal{S}\setminus (\{ s_2 \} \cup \mathcal{S}_1)$, we obtain $\langle s, s_2\rangle < \langle s_2, s_2\rangle$ using the same chain of inequalities of as in \cref{eq:aux_proof_conv} and the inequality $\| s \| \leq \| s_2 \|$ implied by the definition of $\mathcal{S}_2$. We then apply \Cref{lem:verticesSelfMax} with $s_0 = s_2$ and $\mathcal{S}' = \mathcal{S} \setminus ( \{ s_2\} \cup \mathcal{S}_1)$ to find $\delta_2 \in (0,\delta_1]$ such that
\begin{equation}
    \label{eq:z_ownMax} 
    \langle y, x \rangle < \langle x, x \rangle
\end{equation}
for all $x\in B_{\delta_2} (s_2)$ and $y \in \bigcup_{s\in \mathcal{S}'} B_{\delta_2} (s)$. And again by \Cref{lem:setSattracting}, we know that there exists a time $t_2\in \N$ such that $\zz{i}{t} \in \mathcal{S}_{\delta_2}$ for all $t \geq t_2$ and all $i\in [n]$. Combining \cref{eq:z_ownMax} with the model \cref{eq:transformer} implies that a token $\zz{i}{t} \in B_{\delta_2}(s_2)$ is a convex combination of itself and tokens $\zz{j}{t} \in \bigcup_{s\in \mathcal{S}_1} B_{\delta_2}(s) \cup B_{\delta_2} (s_2)$, that is,
\begin{equation}
    \zz{i}{t} \in \mathcal{S}_{\delta_2} \cap \co \left( \bigcup_{s\in \mathcal{S}_1} B_{\delta_2}(s) \cup B_{\delta_2} (s_2) \right) = \bigcup_{s\in \mathcal{S}_1} B_{\delta_2}(s) \cup B_{\delta_2} (s_2) 
\end{equation}
for all $t\geq t_2$. This allows only two possibilities:
\begin{enumerate}[(a)]
    \item $\zz{i}{t} \in B_{\delta_2} (s_2)$ for all $t\geq t_2$.
    \item $\zz{i}{t^\star}\in B_{\delta_2} (s_1)$ for $t^\star > t_2$ and $s_1\in\mathcal{S}_1$. Since $B_{\delta_2} (s_1) \subseteq B_{\delta_1} (s_1)$, then $\zz{i}{t} \in B_{\delta_2} (s_1)$ for all $t\geq t^\star$.
\end{enumerate}
The argument can be applied to every point in $\mathcal{S}_2$, resulting in a collection of radii $\delta_2^1, \dots, \delta_2^{|\mathcal{S}_2|}$ and times $t_2^1, \dots, t_2^{|\mathcal{S}_2|}$. As before, we abuse notation by using $\delta_2$ to denote the smallest of all such radii and $t_2$ to denote the maximum of all such times.

We quickly sketch the argument that finishes the proof by induction. For $N > 2$, we consider the set $\mathcal{S}_N$ of points in $\mathcal{S}$ but not in $\mathcal{S}_1 \cup \dots \cup \mathcal{S}_{N-1}$ with maximum norm. The same arguments that proved the case $N=2$ will conclude that, for all $s_N \in \mathcal{S}_N$, there exists a time $t_N \in \N$ and a $\delta_N > 0$ such that all tokens in $B_{\delta_N}(s_N)$ either (a) stay in $B_{\delta_N}(s_N)$ for all times $t\geq t_N$, or (b) jump to a neighborhood of $s_\ell\in \mathcal{S}_{\ell}$ for some $\ell \in [N-1]$, for which we also have the result by the induction hypothesis.
Let $\mathcal{S}_{M + 1}$ for $M\geq 1$ be the first empty set in the collection $\{ \mathcal{S}_\ell\}_{\ell = 1}^{M + 1}$. Then, setting $\delta = \delta_M$ finishes the proof. A visual sketch of the proof is represented in \Cref{fig:sketchConvergenceSi}.
\end{proof}

\section{Characterization of cluster points via leaders}
\label{sec:leaders}
In this section, we show the results implying parts $(ii)$ and $(iii)$ of \Cref{thm:emergenceClusters}, which characterize the elements of the cluster set $\mathcal{S}$ via special tokens we call leaders. Recall that a token $z_i$ is a leader if there exists time $t\in \N$ such that $\CC{i}{t} = \{ i \}$.
\subsection{Basic properties of leaders}
\label{subsec:basicPropLeaders}
\begin{figure}
\centering
\begin{subfigure}{.45\textwidth}
  \centering
  \includegraphics{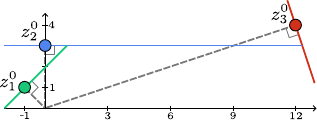}
  \caption{}
\end{subfigure}
\begin{subfigure}{.45\textwidth}
  \centering
  \includegraphics{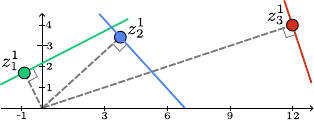}
  \caption{}
\end{subfigure}
\caption{Token values in \cref{ex:becomeLeader} for (a) initial time $t=0$ and (b) time $t = 1$. After one iteration, the token $z_2$ has moved enough to be below the hyperplane separating the tokens influencing token $z_1$ (in green), which causes token $z_1$ to satisfy the leader condition $\CC{1}{1} = \{ 1 \}$ at time $t = 1$.}
\label{fig:becomeLeader}
\end{figure}

We begin proving that leaders exist, they remain invariant for \cref{eq:transformer}, and once they satisfy the leader condition $\CC{i}{t} = \{ i \}$ at time $t\in \N$, they do for all future times.
\begin{lemma}
    \label{lem:BasicPropertiesLeaders}
    The following statements about the set of leaders $\mathcal{L}$ hold:
    \begin{enumerate}[(i)]
        \item $\mathcal{L}$ is not empty.
        \item If $z_i \in \mathcal{L}$ and $t\in\N$ such that $\CC{i}{t} = \{ i \}$, then $\zz{i}{t} = \zz{i}{\ell}$ and $\CC{i}{\ell} = \{ i \}$ for all times $\ell \geq t$.
    \end{enumerate}
\end{lemma}
\begin{proof}
To prove $(i)$, choose a token $z_i$ such that $\| z_i^0 \| = \max_{\ell \in [n]} \| \zz{\ell}{0} \|$. Then, for all $j\neq i$, $\| z_j^0 \| \leq \| z_i^0 \|$. Arguing as in \cref{eq:aux_proof_conv} shows that $\langle z_j^0, z_i^0\rangle < \langle z_i^0, z_i^0 \rangle$, which implies that $\CC{i}{0}  = \{i\}$ and then $z_i\in \mathcal{L}$.

For the proof of $(ii)$, it is enough to prove the result for $\ell = t + 1$, since the argument can be iterated. By choice of $t$, $\CC{i}{t} =\{ i \}$. This can be combined with the model \cref{eq:transformer} to obtain $\zz{i}{t+1}=\zz{i}{t}+{\alpha}/(1+\alpha) \left(\zz{i}{t}-\zz{i}{t}\right) = \zz{i}{t}$. Additionally, $\zz{i}{t}$ satisfies by definition
$\langle \zz{i}{t}, \zz{r}{t} \rangle < \langle \zz{i}{t}, \zz{i}{t} \rangle$ for all $r \neq i$, so we obtain
\begin{multline}
     \label{eq:proofCi1}
     \langle \zz{i}{t+1}, \zz{r}{t+1}\rangle = 
     \frac{1}{1 + \alpha} \langle \zz{i}{t},\zz{r}{t}\rangle +\frac{\alpha}{1 + \alpha}\frac{1}{|\CC{r}{t}|} \sum_{j \in \CC{r}{t}} \langle \zz{i}{t},\zz{j}{t}\rangle \\
    \leq \frac{1}{1 + \alpha} \langle \zz{i}{t},\zz{r}{t}\rangle +\frac{\alpha}{1 + \alpha} \langle \zz{i}{t}, \zz{i}{t} \rangle <
     \frac{1}{1 + \alpha} \langle \zz{i}{t}, \zz{i}{t} \rangle 
     +
     \frac{\alpha}{1 + \alpha} \langle \zz{i}{t}, \zz{i}{t} \rangle = \langle \zz{i}{t+1}, \zz{i}{t+1} \rangle.
\end{multline}
This implies $\CC{i}{t+1} = \{ i \}$, as desired.
\end{proof}
In the proof of \Cref{lem:BasicPropertiesLeaders}(i), we see that tokens with maximum norm satisfy the leader condition initially. However, not all leaders are necessarily determined by their initial values. Indeed, as the next example shows, a token $z_i$ which does not satisfy the leader condition initially, i.e. $\CC{i}{0} \neq \{i \}$, can satisfy it after some time $t>0$.
\begin{example}
\label{ex:becomeLeader}
    Consider the tokens $\zz{1}{}, \zz{2}{}, \zz{3}{} \in \R^2$ with initial values $z_1^0 = (-1,1)^\top$, $z_2^0 = (0,3)^\top$, $z_3^0 = (12,4)^\top$, and fix the step-size parameter $\alpha = 0.5$. In this case, one checks that $\CC{1}{0} = \{2\}$, $\CC{2}{0} = \{3\}$ and $\CC{3}{0} = \{3\}$, so only $z_3$ is determined as a leader at time $t = 0$. We apply the model \cref{eq:transformer} and obtain the values $z_1^1 = (-{2}/{3},{5}/{3})^\top, z_2^1 = (4,{10}/{3})^\top, z_3^1 = (12,4)^\top$. This implies that $\CC{1}{1} = \{1\}$, $\CC{2}{1} = \{3\}$ and $\CC{3}{1} = \{3\}$, so both $z_1$ and $z_3$ are determined as leaders at time $t = 1$. We include an illustration in \Cref{fig:becomeLeader}.
\end{example}
\subsection{Correspondence between leaders and vertices}
\label{subsec:vertices=leaders}
We now prove that a token $\zz{i}{}$ is a leader if and only if its limiting value is a vertex of $\mathcal{K}$. This will follow from the next two lemmas, each of which establishes one direction of the implication.
\begin{lemma}
\label{lem:ownMaxImpliesVertex}
    If token $\zz{i}{}$ is a leader, it converges in finite time to a vertex of $\mathcal{K}$. 
\end{lemma}
\begin{proof}
First, we show that there exists a time $t^\star\in\N$ such that $z_i^t \in \partial \mathcal{K}$ for all $t\geq t^\star$. By \Cref{lem:BasicPropertiesLeaders}(ii), there exists a time $t^\star \in \N$ such that $z_i^t = z_i^{t^\star}$ for all $t\geq t^\star$. This, combined with \Cref{lem:setSattracting}, implies $\zz{i}{t}\in \mathcal{S}$. Finally, \Cref{lem:characterisationS} ensures $\mathcal{S} \subset \partial \mathcal{K}$, and so $\zz{i}{t}\in \partial\mathcal{K}$.

Now, assume by contradiction that $z_i^t \notin \mathcal{V}$. Since $z_i^t\in \partial \mathcal{K}$, it lies in a face spanned by vertices $v_1, \dots, v_r \in \mathcal{V}$ for $r\geq 2$, which allows us to write $z_i^t = \sum_{j = 1}^r \beta_j v_j$, where $\sum_{j=1}^r \beta_j = 1$ and  $\beta_j \in (0,1)$. Now, for any $j\in [r]$,
\begin{equation}
\label{eq:aux_leadersAreVertices}
    \langle v_j - z_i^t, z_i^t \rangle \leq \max_{v\in \mathcal{V}} \langle v -z_i^t, z_i^t \rangle = \max_{z\in \mathcal{K}} \langle  z - z_i^t, z_i^t \rangle \leq \max_{z\in  \co(Z^t)} \langle  z - z_i^t, z_i^t \rangle = 0,
\end{equation}
where in the last equality we use $\CC{i}{t} = \{ i \}$. We now claim that the inequality is in fact an equality. To see why, note that if there exists $j\in [r]$ such that $\langle v_j -z_i^t, z_i^t \rangle < 0$, then we would obtain the contradiction
\begin{equation}
    0 > \sum_{j=1}^r \beta_j \langle v_j - z_i^t, z_i^t \rangle = \Bigg< \sum_{j=1}^r \beta_j v_j - \sum_{j=1}^r \beta_j z_i^t, z_i^t \Bigg> = \langle z_i^t - z_i^t, z_i^t \rangle = 0.
\end{equation}
To conclude, we define the closed half-plane $\mathcal{H} = \{z \in \R^d \, \, : \,\, \langle z - z_i^t, z_i^t \rangle \geq 0\}$. Using $\mathcal{K} \subseteq \co (\ZZ^t)$ and the assumption $\CC{i}{t} = \{ i \}$, we obtain $\mathcal{K} \cap \mathcal{H}\subseteq \co (\ZZ^t) \cap \mathcal{H} = \{ z_i^t\}$. But we have just proved that $\{v_1, \dots, v_r\} \subset \mathcal{K} \cap \mathcal{H}$ by showing that \cref{eq:aux_leadersAreVertices} is an equality. This implies $z_i^t = v_j$ for $j\in [r]$, contradicting the hypothesis $z_i^t \notin \mathcal{V}$.
\end{proof}
\begin{lemma}
\label{lem:vertexImpliesOwnMax}
    If $v$ is a vertex of $\mathcal{K}$, then $v = \zz{i}{t}$ for some leader $z_i$ and time $t\in \N$.
\end{lemma}
\begin{proof}
First, we show that $\langle s, v\rangle \leq \langle v, v \rangle$ for all $s\in \mathcal{S}$. Because $s\in \partial\mathcal{K}$, it can be written as a convex combination of the vertices of $\mathcal{K}$, that is, $s = \sum_{j = 1}^m \beta_j v_j$ for some $\beta_j \in [0,1]$ such that $\sum_{j = 1}^m \beta_j = 1$. Then, we obtain
\begin{equation}
    \langle s,v \rangle = \Big< \sum_{j = 1}^m \beta_j v_j , v\Big> = \sum_{j = 1}^m \beta_j \langle v_j, v \rangle \leq \sum_{j = 1}^m \beta_j \langle v, v \rangle = \langle v, v \rangle,
\end{equation}
where in the inequality we have used that $v\in \mathcal{S}$ by \Cref{lem:VinS}, and thus $\langle v_j, v \rangle \leq \langle v, v \rangle$ for all $j\in [m]$. We consider two cases, one in which the inequality $\langle s, v\rangle < \langle v, v \rangle$ is strict for all $s\in \mathcal{S} \setminus \{ v\}$, and another one in which it is not.

\subsection*{%
\texorpdfstring{Case 1: $\langle s, v \rangle < \langle v, v \rangle, \; \forall s \in \mathcal{S} \setminus \{v \}$}
{Case 1: <s,v> < <v,v>, for all s in S \ v}%
}

The hypothesis of \Cref{lem:verticesSelfMax} hold with $s_0 = v$ and $\mathcal{S}' = \mathcal{S}\setminus \{ v \}$. Then, there exists $\delta>0$ such that
\begin{equation}
    \label{eq:ownmax3} 
    \langle y, x \rangle < \langle x, x \rangle
\end{equation}
for all $x\in B_{\delta} (v)$ and $y \in \bigcup_{s\in \mathcal{S}'} B_{\delta} (s)$.
Notice that $\delta$ can be assumed small enough so that \Cref{th:noCirculation} applies, so for every $i \in [n]$, there exists $s_i \in \mathcal{S}$ and $t_i \in \mathbb{N}$ such that $\zz{i}{t} \in B_\delta(s_i)$ for all $t\geq t_i$. In particular, it is also true for all $t\geq t^\star$, where $t^\star$ denotes the maximum of all such times $t_i$. Then, \cref{eq:ownmax3} implies
\begin{equation}
    \label{eq:V_aux}
    \langle \zz{j}{t},\zz{i}{t} \rangle < \langle \zz{i}{t},\zz{i}{t} \rangle
\end{equation}
for all $\zz{i}{t} \in B_{\delta}(v)$ and all $\zz{j}{t}\notin B_{\delta}(v)$. Choose $z_M^t \in B_{\delta}(v)$ such that $\| z_M^t \| = \max_{z_\ell^t \in B_\delta(v)} \| \zz{\ell}{t}\|$. The argument used in  \cref{eq:aux_proof_conv} implies that $\langle \zz{j}{t}, z_M^t \rangle < \langle z_M^t, z_M^t \rangle$ for all $\zz{j}{t} \in  B_\delta(v) \setminus \{z_M^t\}$. This is combined with \cref{eq:V_aux} to obtain $\CC{M}{t} = \{ M \}$ and thus $z_M$ is a leader. By \Cref{lem:ownMaxImpliesVertex}, $z_M$ must be a vertex and so $v = z_M$.

\subsection*{%
\texorpdfstring{Case 2: $\langle s, v \rangle = \langle v, v \rangle,\; \forall s\in\mathcal{S}^\star\subseteq \mathcal{S} \setminus \{ v\},\, \mathcal{S}^\star \neq \emptyset$}
{Case 2: <s,v> = <v,v>, for all s in S* subset of S minus v, S* nonempty}%
}

By the same arguments used in Case 1 with $\mathcal{S}' = \mathcal{S}\setminus (\{ v \} \cup \mathcal{S}^\star)$, there exist $\delta_1>0$, $s_i\in \mathcal{S}$, and a time $t^\star\in \N$ such that $\zz{i}{t} \in B_{\delta_1}(s_i)$ for all $t \geq t^\star$ and all $i\in [n]$, $B_{\delta_1} (s') \cap B_{\delta_1} (s'') = \emptyset$ if $s' \neq s''$, and
\begin{equation}
    \label{eq:aux_case2}
    \langle \zz{j}{t},\zz{i}{t} \rangle < \langle \zz{i}{t},\zz{i}{t} \rangle
\end{equation}
for all $\zz{i}{t} \in B_{\delta_1}(v)$ and all $\zz{j}{t} \in \bigcup_{s\in \mathcal{S}'} B_{\delta_1}(s)$. Note that the same is true if one replaces $\delta_1$ for any $\delta \in (0, \delta_1]$. If we denote $\mathcal{S}^\star = \{ s_1^\star, \dots, s_L^\star \}$, then \cref{eq:aux_case2} implies that $\zz{i}{t} \in B_\delta (v)$ is attracted to $N_0$ tokens in $B_\delta (v)$ and $N_j$ tokens in $B_\delta (s_j^\star)$ for $j \in [L]$. By construction, $N_j \in \N$ and $\sum_{j=0}^L N_j = |\CC{i}{t}|$.

When $N_0 = | \CC{i}{t}|$, the arguments of Case 1 end the proof, since among the tokens in $N_0$, one with maximum norm $\zz{M}{}$ can be picked. Therefore, we focus on the case $N_0 \leq | \CC{i}{t} | - 1$. We use the notation $s_0^\star = v$ to simplify the following calculations. Define $\Bar{v} = {\alpha}/(1 + \alpha)|\CC{i}{t}|^{-1} \sum_{j=0}^L N_j ( s_j^\star - s_0^\star)$. We now claim (and will prove below) that there exists a constant $c > 0$, depending only on $\alpha$, $n$ and $s_0^\star, s_1^\star, \dots, s_L^\star$, such that $\| \Bar{v} \| \geq c$. In particular, $c$ is independent of $\delta \in (0 , \delta_1]$ used to identify the attracting tokens. We may then assume that $\delta < c/4$ and show that the assumptions in Case 2 lead to a contradiction. We use the dynamics \cref{eq:transformer} to define $\Bar{w} \coloneqq \zz{i}{t+1} - \zz{i}{t} = {\alpha}/(1+\alpha) |\CC{i}{t}|^{-1} \sum_{j \in |\CC{i}{t}|}(\zz{j}{t} - \zz{i}{t})$. 
Recall that, by the choice of $\delta$ and of the time $t$, we have $\zz{i}{t} \in B_{\delta}(s_0^\star)$. Moreover, for every token $\zz{\ell}{}$ we can find $j\in [L]$ such that $\zz{\ell}{t} \in B_\delta (s_j^\star)$, and there are exactly $N_j$ tokens inside the ball $B_\delta (s_j^\star)$. Using these observations, we can estimate
\begin{align}
\label{eq:>c_2}
    \| \Bar{v} - \Bar{w} \| &= \frac{\alpha}{1+\alpha} \Bigg\| \zz{i}{t} - s_0^\star  +  \frac{1}{|\CC{i}{t}|} \sum_{j=0}^{L}  N_j s_j^\star - \frac{1}{|\CC{i}{t}|} \sum_{j=0}^{L} \sum_{\zz{\ell}{t} \in B_\delta (s_j^\star)} \zz{\ell}{t}    \Bigg\| \\ 
    &<  \| \zz{i}{t} - s_0^\star \| +  \frac{1}{|\CC{i}{t}|} \sum_{j=0}^{L} \sum_{\zz{\ell}{t} \in B_\delta (s_j^\star) } \| s_j^\star - \zz{\ell}{t} \| \leq 2\delta, \nonumber
\end{align}
where in the first inequality we use $\alpha / (1 + \alpha) < 1$ and the triangle inequality, whereas in the second inequality we use that $\zz{i}{t} \in B_{\delta}(s_0^\star)$ and $\zz{\ell}{t} \in B_\delta (s_j^\star)$ for all $j$. Finally, recalling our claim that $\| \Bar{v} \| \geq c$ and that we have chosen $\delta < c / 4$, we obtain from \cref{eq:>c_2} that $\| \Bar{w} \| \geq \| \Bar{v} \| - \| \Bar{v} - \Bar{w} \| \geq c - 2\delta > {c}/{2}$. But $z_i^t \in B_\delta(s_0^\star)$ for $t\geq t^\star$, yielding the contradiction $\| \Bar{w} \|  =  \| \zz{i}{t+1} - \zz{i}{t} \| \leq \| \zz{i}{t+1} - s_0^\star \| + \| s_0^\star - \zz{i}{t} \| \leq 2\delta < {c}/{2}$.

To finish the proof, we need to verify the claim that $\| \Bar{v} \| \geq c$ where $c > 0$ depends only on $\alpha$, $n$ and $s_0^\star, s_1^\star, \dots, s_L^\star$. From the definition of $\Bar{v}$, we obtain
\begin{align}
\label{eq:claim1}
    \frac{|\CC{i}{t}|^2 (1 + \alpha)^2}{\alpha^2} \| \Bar{v}\|^2 &= \sum_{i,j=1}^L N_i N_j \langle s_j^\star - s_0^\star, s_i^\star - s_0^\star \rangle \\
    &\geq \min_{N_1, \dots, N_L} \sum_{i,j=1}^L N_i N_j \langle s_j^\star - s_0^\star, s_i^\star - s_0^\star \rangle \eqqcolon \rho, \nonumber
\end{align}
where we minimize over integers $N_1, \dots, N_L$ satisfying $0 \leq N_j \leq n$ for all $j \in [L]$ and $\sum_{j=1}^L N_j = |\CC{i}{t}| - N_0 \geq 1$. Notice that the minimum value $\rho$ is nonnegative, is attained by optimal choices $N_1^\star,\dots, N_L^\star$, and depends only on the points $s_0^\star, s_1^\star,\dots, s_L^\star$, which are given. Reordering \cref{eq:claim1} and using the inequality $|\CC{i}{t}| \leq n$ gives
\begin{equation}
    \| \Bar{v} \| \geq \frac{{}\sqrt{\rho} \alpha}{(1 + \alpha) n} \eqqcolon c \geq 0,
\end{equation}
where $c$ depends only on $\alpha$, $n$, and the points $s_0^\star,s_1^\star, \dots, s_L^\star$. We prove next that $\rho > 0$, and consequently $c > 0$. Indeed, assume by contradiction that $\rho = 0$. In this case, $\sum_{j=1}^L N_j^\star (s_j^\star - s_0^\star) = 0$, which implies $s_0^\star = \sum_{j=1}^L ( {N_j^\star} / {\sum_{k=1}^L N_k^\star}) s_j^\star = \sum_{j=1}^L \gamma_j s_j^\star$, where $\gamma_j \in (0,1)$ and $\sum_{j=1}^L \gamma_j = 1$. Thus, $s_0^\star$ is a convex combination of $s_1^\star,\dots, s_L^\star \in \mathcal{K}$. This contradicts that $s_0^\star = v$ is a vertex of $\mathcal{K}$ and, as such, an extreme point.
\end{proof}
\section{The role of clustering in sentiment analysis}
\label{sec:numerics}
In this section, we use our clustering results to design an interpretable transformer-based model to solve the supervised learning task of sentiment analysis, that asks to classify movie reviews according to their positive or negative assessment of the movie. 
Despite its simplicity, this is a relevant task in many text-based applications such as marketing and publicity, where transformers have achieved state-of-the-art performance~\cite{acheampong2021transformer}. The proposed model contains only three components with explainable roles. The first is an encoder, which maps words in a movie review to tokens in $\R^d$ and whose role is to select meaningful words as leaders. The second and core component is our transformer \cref{eq:transformer}, whose role is to capture `context' by clustering tokens around these leaders. The third and final component is a decoder, which projects the token values after the last transformer layer to a positive or negative sentiment prediction. We train our model to classify a range of sample movie reviews and confirm our interpretation with examples, showing that leaders indeed correspond to meaningful words and that our transformer captures `context' by clustering most words around the leaders. Our code is available from \url{https://github.com/DCN-FAU-AvH/clustering-hardmax-transformers}.

\subsection{The task} 
\label{ss:sentiment_task}
\begin{figure}
    \centering
    \includegraphics[width = 0.5\textwidth]{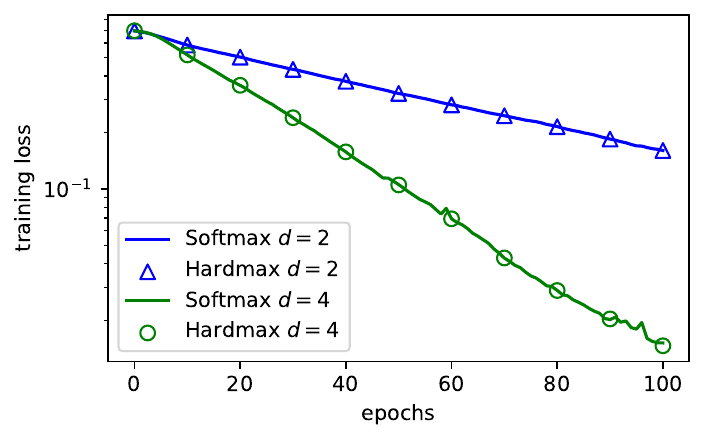}
    \caption{Loss on the training set for encoder dimensions $d \in \{2, 4\}$, calculated at every epoch with the regularized softmax model used for training, and every 10 epochs with our hardmax model.}
    \label{fig:exp_sm_hm}
\end{figure}
For our simulations, we focus on predicting the sentiment of movie reviews of the benchmark IMDb dataset ~\cite{maas2011}, consisting of $50\,000$ movie reviews labeled as positive (1) or negative (0). We fix the length of each review to be $n = 128$ words; the $16\,914$ reviews longer than this are truncated, while the $33\,086$ reviews that are shorter are padded with the meaningless word \texttt{<PAD>}~\cite{dwarampudi2019padding}.
Each of the $W = 121\,296$ distinct words in our dataset is identified with a canonical basis vector in $\R^W$. Thus, a movie review $X$ is an $n \times W$ matrix whose rows are canonical basis vectors. 
\subsection{Model and training}
Fix a transformer depth $K\in \N$ and an encoder dimension $d \in \N$. Given a movie review $X \in \R^{n \times W}$ and its label $y \in \{0 , 1  \}$, our transformer model outputs a value $\hat{y} \in [0,1]$ by performing the following operations:
\begin{enumerate}
    \item Encoder: set $Z^0 = XE$ for a trainable encoder matrix $E\in \R^{W \times d}$.
    \item Transformer: compute $Z^K = \mathcal{T}(Z^0)$ where $\mathcal{T}$ is our transformer \cref{eq:transformer} with fixed $A = I\in \R^{d \times d}$ and trainable step-size $\alpha > 0$.
    \item Decoder: compute the average token $\Bar{z} = \frac{1}{n} (Z^K)^\top \mathbf{1}$ and the value $\hat{y} = \sigma(\Bar{z}^\top w \, + \, v)$, where $\mathbf{1} \in \R^n$ is the vector with entries equal to one and $\sigma(x) = {1}/({1 + {\rm e}^{-x})}$ is the sigmoid function. The decoder vector $w\in \R^d$ and bias $v\in \R$ are trainable parameters.
\end{enumerate}
The value $\hat{y} \in [0,1]$ is rounded to the nearest integer to obtain a prediction in the discrete set $\{ 0,1\}$. We train the parameters $E \in \R^{W \times d}$, $w \in \R^d$, $v \in \R$, and $\alpha > 0$ for a fixed  transformer depth $K=8$ using \texttt{PyTorch} \cite{paszke2019pytorch}. Since automatic differentiation capabilities of this toolbox require the model to be differentiable with respect to its parameters, but our transformer model is not because the hardmax function is not continuous, we simplify the implementation by replacing the hardmax with its differentiable softmax approximation in \cref{e:att-softmax} with $\tau = 0.001$. We then use the gradient-based algorithm Adam \cite{kingma2014adam} to minimize the mean binary cross-entropy loss
\begin{equation}
    \frac{1}{N} \sum_{i=1}^N - \left(y_i \log (\hat{y}_i)  + (1 - y_i) \log (1 - \hat{y}_i)\right)
    \label{eq:loss-experiment}
\end{equation}
calculated using a \textit{training set} of $N=35\,000$ reviews. The remaining reviews form a \textit{test set} used for validation purposes. We use a learning rate of $0.001$, a batch size of $64$, and train for $100$ epochs, resulting in approximately $55\,000$ gradient descent steps in total.
Note that, as shown in \Cref{fig:exp_sm_hm}, computing the training loss in \cref{eq:loss-experiment} for the softmax and hardmax models returns almost identical values, confirming that the former is a good approximation of the latter.

Finally, we stress that fixing $A = I$ in our transformer, rather than training $A$ over the set of symmetric positive definite matrices, is not a restriction. Indeed, as pointed out at the start of \Cref{sec:problemFormulation}, it amounts to considering the dynamics of transformed tokens $\Tilde{z}_i = B \Tilde{z}$ where $B\in \R^{d\times d}$ is any invertible matrix satisfying $A = B^\top B$, which can be `absorbed' into a transformed encoder matrix $\Tilde{E} = E B^\top$.
Moreover, fixing $A = I$ simplifies the training process by reducing the number of parameters and, more importantly, avoiding the need to enforce the positive definiteness constraint.

\subsection{Results}
\label{ss:numericalResults}
\begin{figure}
  \centering
  \begin{subfigure}[b]{0.45\textwidth}
    \includegraphics[width=\textwidth]{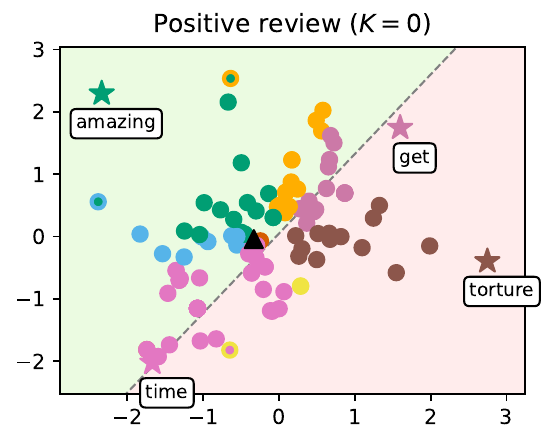}
    \label{fig:1}
  \end{subfigure}
\vspace{0.5em}  
  \begin{subfigure}[b]{0.45\textwidth}
    \includegraphics[width=\textwidth]{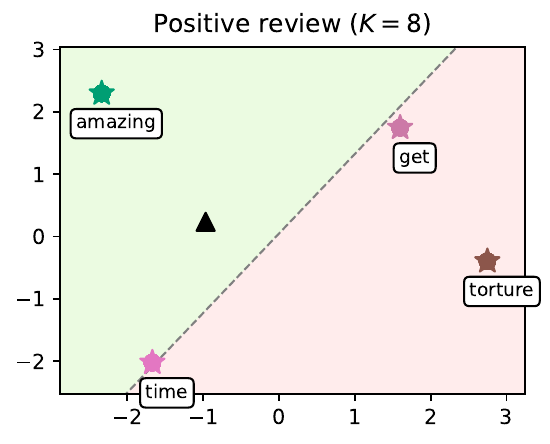}
    \label{fig:2}
  \end{subfigure}
    \begin{subfigure}[b]{0.45\textwidth}
  \includegraphics[width=\textwidth]{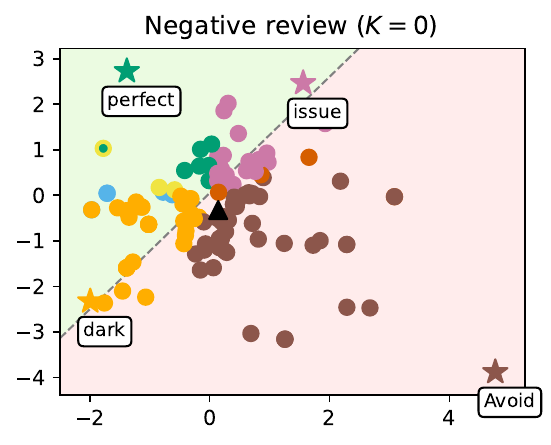}
    \label{fig:3}
  \end{subfigure}
  \begin{subfigure}[b]{0.45\textwidth}
    \includegraphics[width=\textwidth]{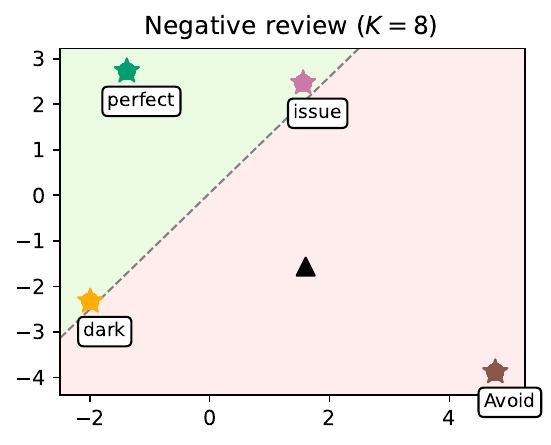}
    \label{fig:4}
  \end{subfigure}
  \caption{Evolution of the words of a positive review (top) and a negative review (bottom), as they are processed by the transformer layers. Color coded as \Cref{fig:dependence_IC}. Leaders are represented with stars and tagged with the word they encode, circles are the remaining tokens, and the black triangle is the average token. The dashed line represents the hyperplane $\mathcal{H}(z) = z^\top w \, + \, v = 0$, which separates the half-plane identified with the positive class (shaded in green) from the one identified with the negative class (shaded in red).}
  \label{fig:expRev}
\end{figure}
\begin{figure}
\centering
\includegraphics[width=0.6\textwidth]{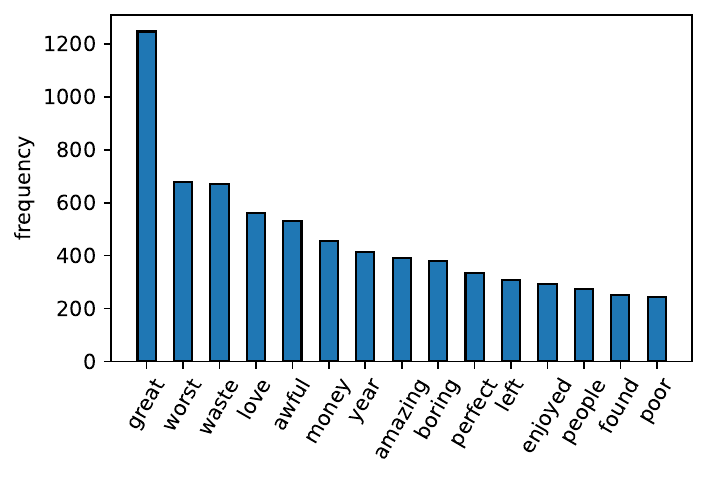}
\caption{Histogram of the 15 most frequent leaders $z_i$ of correctly classified test reviews and satisfying $|\mathcal{H}(z_i)| \geq 2$, meaning that they are situated far from the separating hyperplane $\mathcal{H}$.}
\label{fig:freqleaders}
\end{figure}

We confirm that leaders are meaningful words with two illustrative examples in \Cref{fig:expRev}, where we fix the encoder dimension to $d = 2$ for visualization purposes and plot the initial and final token values. The two selected reviews belong to the test set and are correctly classified. Notice that $K = 8$ transformer layers are sufficient to approximate the asymptotic state $(t = \infty)$ with all tokens clustered around the leaders. We observe that amongst the leader words, those conveying sentiment (\texttt{amazing}, \texttt{torture}, \texttt{perfect}, \texttt{Avoid}) are placed the furthest away from the hyperplane $\mathcal{H}(z) = z^\top w \, + \, v = 0$ separating the positive and negative half-planes. This causes the average final value of the tokens, understood as the average sentiment of the review and represented with a black triangle, to be moved towards the half-plane identified with the correct label. We further confirm our interpretation in \Cref{fig:freqleaders}, where we show the 15 most frequent leaders in correctly classified test reviews which satisfy $|\mathcal{H}(z)| \geq 2$, so they are situated far from the separating hyperplane $\mathcal{H}$. These leaders, which are the most influential when calculating a prediction based on the average token, are observed to mostly encode words related with sentiment.

Next, we explore the influence of the encoder dimension by training our model for $d \in \{2, 4, 8, 16 \}$. The proportion of correctly classified movie reviews in the test set grows from around $80\%$ for $d=2$ to around $90\%$ for $d = 16$. This phenomenon can be attributed to two main reasons. First, increasing the encoder dimension results in more training parameters, providing more expressive power to the model. Second, a higher encoder space allows for clustering patterns to become more complex and, thus, capture `context' more accurately. A good measure for this is the number of leaders determined up to layer $K$. \Cref{tab:numberLeaders} shows that the average number of leaders per review grows with the encoder dimension $d$. Finally, we measure the proportion of leaders determined from the initial token values, and verify that such proportion is extremely close to one for all encoder dimensions considered in our study. This indicates that, even though the existence of leaders not determined initially is theoretically possible (see \cref{ex:becomeLeader}), this phenomenon is rarely observed in this application.
\begin{table}
\centering
\begin{tabular}{cccccc} \toprule
    \thead{$d$} & \thead{$\overline{m}$} &  \thead{$\sigma_m$} &  \thead{$m_{\min}$} & \thead{$m_{\max}$} & \thead{${\overline{m_0 / m}}$} \\ \midrule
    2  & 5.960 & 1.247 & 2 & 11 & 0.995 \\
    4  & 19.321 & 4.095 & 4 & 34 & 0.999  \\
    8  & 49.866 & 13.131 & 6 & 84 & 0.999 \\
    16 & 75.423 & 23.646 & 7 &  113 & 1.000 \\
 \bottomrule
\end{tabular}
\caption{Statistical information on leaders in the test set. For each encoder dimension $d \in \{ 2, 4, 8, 16 \}$, average number of leaders ($\overline{m}$), standard deviation of the number of leaders ($\sigma_m$), minimum number of leaders ($m_{\min}$), maximum number of leaders ($m_{\max}$) and average proportion of leaders determined initially ($\overline{m_0 / m}$).}
\label{tab:numberLeaders}
\end{table}

\section{Conclusions and perspectives}
\label{sec:conclusion}
In this paper, we studied a class of transformers called pure-attention hardmax transformers, we characterized their behavior in the infinite-depth limit, and we leveraged this understanding to design an interpretable model for sentiment analysis. Thanks to the geometric nature of our model \cref{eq:transformer}, we proved the existence of a clustered equilibrium and we identified the cluster points via special tokens we called leaders. In language modeling applications, where transformers perform exceptionally well, leaders typically correspond to particularly meaningful words in a sentence or paragraph, which capture the meaning of the text.

On the practical side, we illustrated what our theoretical results mean in a real machine learning application. We built a minimal transformer model and used it to interpret the role of clustering as a way to capture `context' when solving a sentiment analysis problem. With our simulations, we visualized how the model selects relevant words as leaders, and then exploits the clustering phenomenon to `filter out' unimportant words, effectively letting `context' emerge.

The analysis in our work represents a step towards understanding the complex mechanisms behind transformers and, crucially, provides rigorous results that improve their interpretability. However, further study is needed to bridge the gap between models used in practice and models for which rigorous theory exists. We conclude by outlining some outstanding open problems.

\subsection*{Clustering for arbitrary parameter matrices} A first open problem remains to understand whether clustering occurs when the matrices $A, V \in \R^{d\times d}$ do not satisfy \Cref{ass:IC}. Our analysis relies on the bilinear form $(x,y)\mapsto \langle Ax , y \rangle$ being an inner product, which is not the case when $A$ is not symmetric or positive definite. Addressing these difficulties is important because symmetry and positive definiteness usually fail for real-life transformers, where low-rank matrices $A$ are used to ease training. If $A$ is semidefinite, then our clustering results apply to components of the initial tokens onto the positive definite eigenspace, but nothing can be said about the dynamics in the kernel of $A$.
The non-symmetric case is also relevant, as the lack of symmetry arises when $A = Q^\top K$ is trained as the product of `query' and `key' matrices~\cite{vaswaniAttentionAllYou2017}. Additionally, the assumption $V = \alpha I$ is also quite restrictive, as typically $V$ is trained as a dense matrix. 

\begin{figure}
\centering
\begin{subfigure}{.49\linewidth}
  \centering
  \includegraphics[width = 0.95\linewidth]
  {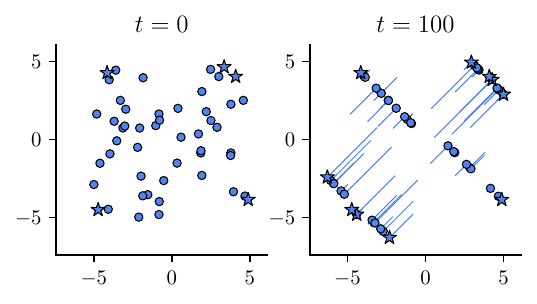}
  \caption{}
  \label{fig:conjV(a)}
\end{subfigure}%
\begin{subfigure}{.49\linewidth}
  \centering
  \includegraphics[width = 0.95\linewidth]
  {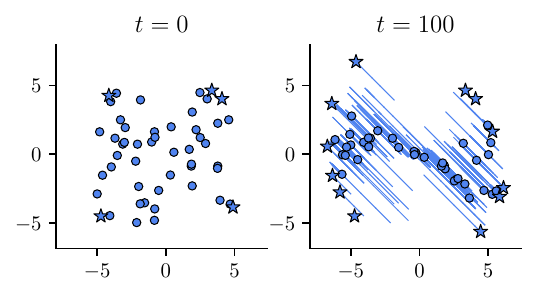}
  \caption{}
  \label{fig:conjV(b)}
\end{subfigure}
\caption{Simulations of \cref{eq:transformer_lowRankV} with $n = 50$ tokens in $\R^2$, $A = I$, and two choices for the vectors $u$, $v$. In panel (a), where  $u=v = (1,1)^\top$, tokens cluster along hyperplanes perpendicular to $v$. In panel (b), where $u = (1/2,-1/2)^\top$ and $v = (1,1)^\top$, tokens appear to reach an equilibrium without clear clusters. Leaders are indicated by stars in all plots.}
\label{fig:conjV}
\end{figure}

As an example of the more complicated behavior that can emerge for general $V$, let us consider the rank-1 case $V = u v^\top$. We assume $u^\top v \neq -1$ to ensure that $I + V$ is invertible. Using the Sherman--Morrison--Woodbury formula to compute its inverse, we can (after some calculations) replace \cref{eq:transformer_a} with
\begin{equation}
\label{eq:transformer_lowRankV}
\zz{i}{t+1}=\zz{i}{t}+\frac{1}{1+ u^\top v}\,\frac{1}{|\CC{i}{t}|}\sum_{j\in \CC{i}{t}} u v^\top\big(\zz{j}{t}-\zz{i}{t}\big).
\end{equation}
We simulate these dynamics in \Cref{fig:conjV} for two choices of $u \in \R^2$ and $v\in \R^2$. In the `symmetric' case $u = v = (1, 1)^\top$ shown in \Cref{fig:conjV}(a), tokens cluster along hyperplanes perpendicular to $v$. This mimics the behavior of `softmax' self-attention models, where particles cluster on at most three hyperplanes \cite{geshkovski2023emergence}, with the key difference that more hyperplanes can arise. In contrast, for a `non-symmetric' case with $u = (1/2,-1/2)^\top$ and $v = (1,1)^\top$ shown in \Cref{fig:conjV}(b), tokens appear to reach an equilibrium without any evident clustering pattern. We therefore wonder how much can be said about the asymptotic behavior of tokens for general matrices $V$. We leave this question to future work, pointing interested readers to \cite{geshkovski2023emergence} for analysis and conjectures with more general $V$ for `softmax' self-attention models.

\subsection*{Interpretability in general transformer models} From a modeling point of view, we have focused on pure-attention transformers with normalization sublayers. It is natural to wonder if our results extend to transformers with feed-forward sublayers, which are more common in applications and appear to be necessary to ensure that transformers have sufficient approximation power~\cite{yunAreTransformersUniversal2020,alberti2023sumformer}. In this regard, a promising direction of study is to understand to which extent the clustering mechanism of self-attention improves the performance of regular feed-forward networks. Another modeling choice has been to focus on single-head self-attention of the form \cref{eq:attentionLayer}. However, full transformers implement a parallel version of self-attention, called \textit{multi-head attention}. Given a \textit{head number} $H \in \N$, the self-attention function $\mathcal{A}(Z)$ in \cref{eq:attentionLayer} is redefined as $\mathcal{A}_i(\ZZ) = \zz{i}{} + \sum_{h=1}^H \sum_{j=1}^n \Lambda_{ij}(\ZZ, A^h) V^h \zz{j}{}$. Our work provides understanding of the so-called \textit{single-head} case $H = 1$, but it is not clear how the asymptotic behavior changes when different heads interact.

It remains to be seen if mathematical analysis of transformers can explain all additional features that these highly complex models have in real-life applications, but we believe it has the potential to help shed light on their inner workings. 
%
%
\section*{Acknowledgements}
The authors thank Borjan Geshkovski for valuable conversations. This work was funded by the Humboldt Research Fellowship for postdoctoral researchers, the Alexander von Humboldt-Professorship program, the European Union (Horizon Europe MSCA project ModConFlex, grant number 101073558), AFOSR Proposal 24IOE027, the COST Action MAT-DYN-NET, the Transregio 154 Project of the DFG, grants PID2020-112617GB-C22 and TED2021-131390B-I00 of MINECO (Spain), and the Madrid Government - UAM Agreement for the Excellence of the University Research Staff in the context of the V PRICIT (Regional Programme of Research and Technological Innovation). 
%
%
\bibliographystyle{abbrvnat}    
\bibliography{refs}

\vfill

\end{document}